\documentclass[10pt,twocolumn,letterpaper]{article}

\usepackage{3dv}
\usepackage{times}
\usepackage{epsfig}
\usepackage{graphicx}
\usepackage{amsmath}
\usepackage{amssymb}

\usepackage{bm}

\usepackage{amsmath,amsthm,amssymb}
\usepackage{graphicx}
\usepackage{textcomp}
\usepackage{wrapfig}
\usepackage{subfig}
\usepackage{color}
\usepackage{xspace}
\usepackage{overpic}
\usepackage{subfig}
\usepackage{wrapfig}
\usepackage{enumitem}
\usepackage{mathrsfs}
\usepackage{multirow}
\usepackage[ruled,vlined]{algorithm2e}

\usepackage{booktabs}
\usepackage{caption}
\definecolor{turquoise}{cmyk}{0.65,0,0.1,0.1}
\definecolor{purple}{rgb}{0.65,0,0.65}
\definecolor{dark_green}{rgb}{0, 0.4, 0}
\definecolor{dark_blue}{rgb}{0, 0, 0.4}
\definecolor{orange}{rgb}{0.6, 0.3, 0.0}
\definecolor{red}{rgb}{0.8, 0.2, 0.2}
\definecolor{brown}{rgb}{0.5, 0.16, 0.16}

\newcommand{\kx}[1]{{\color{black}#1}}

\newcommand{\zq}[1]{{\color{black}#1}}

\newcommand{\revised}[1]{{\color{black}#1}}

\usepackage[ruled]{algorithm2e} 

\usepackage[pagebackref=true,breaklinks=true,letterpaper=true,colorlinks,bookmarks=false]{hyperref}

\threedvfinalcopy 

\def\threedvPaperID{56} 

\ifthreedvfinal\fi
\begin{document}


\title{COALESCE: Component Assembly by Learning to Synthesize Connections}

\author{Kangxue Yin\\
Simon Fraser University\\
\and
Zhiqin Chen\\
Simon Fraser University\\
\and
Siddhartha Chaudhuri\\
Adobe Research, IIT Bombay\\
\and
Matthew Fisher\\
Adobe Research\\
\and
Vladimir G. Kim\\
Adobe Research\\
\and
Hao Zhang\\
Simon Fraser University\\
}

\maketitle
\thispagestyle{empty}


\begin{abstract}
We introduce COALESCE, the first data-driven framework for component-based shape assembly which employs deep learning to synthesize part connections.
To handle geometric and topological mismatches between parts,  we remove the mismatched portions via erosion, and rely on a joint synthesis step, which is learned from data, to fill the gap and arrive at a natural and plausible part joint.
Given a set of input parts extracted from different objects, COALESCE automatically aligns them and synthesizes plausible joints to connect the parts into a coherent 3D object represented by a mesh. 
The joint synthesis network, designed to focus on joint regions,
reconstructs the surface between the parts by predicting an implicit shape representation that agrees with existing parts, while generating a smooth and topologically meaningful connection.
%
%
We demonstrate that our method significantly outperforms prior approaches including baseline deep models for 3D shape synthesis, as well as state-of-the-art methods for shape completion.

\end{abstract}

\section{Introduction}
\label{sec:intro}

\begin{figure}
  \centering
  \includegraphics[width=0.95\linewidth]{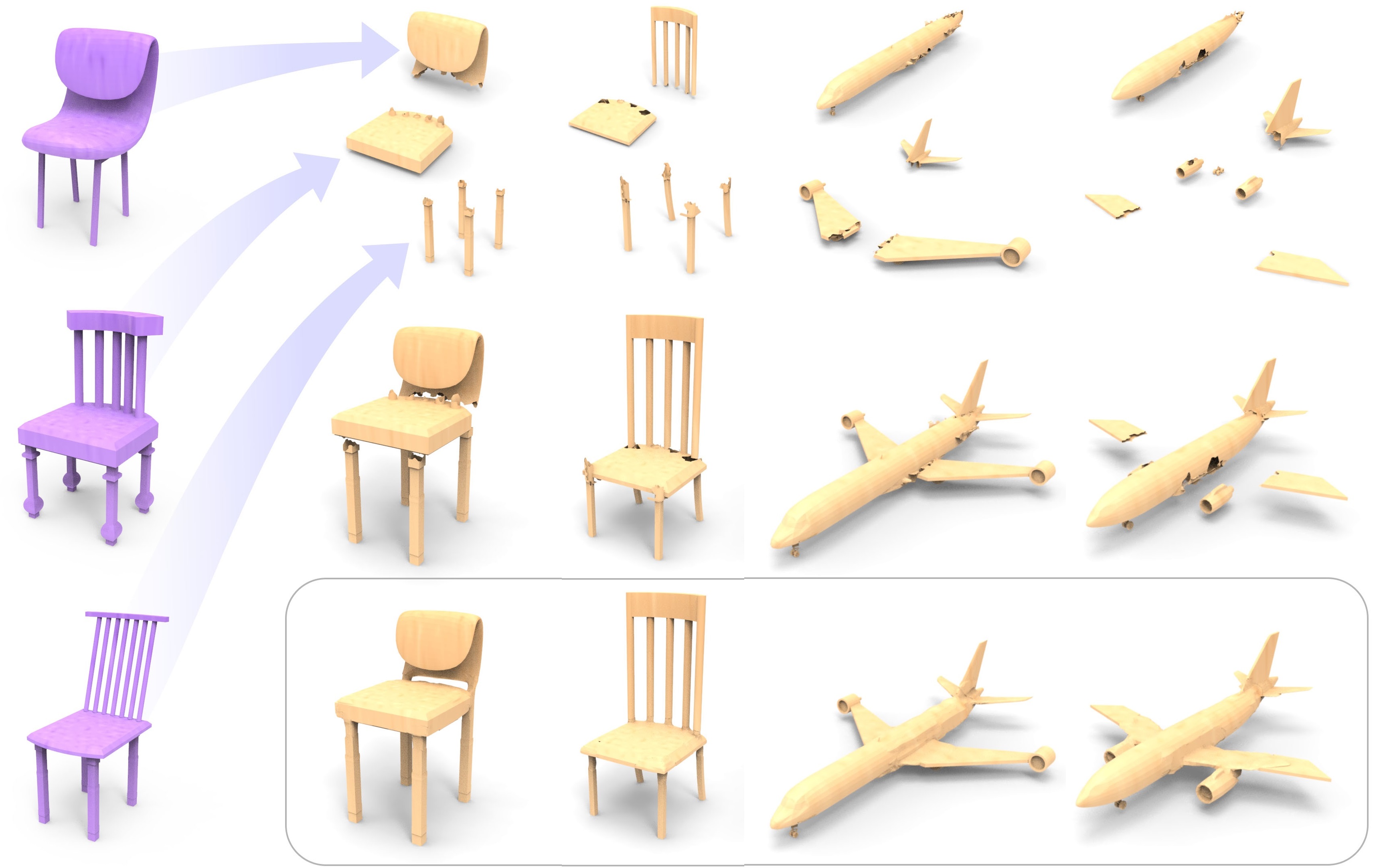}
	\caption{
	Our deep learning framework, COALESCE, assembles 3D shapes by a learning-based approach to synthesize part connections. Given a set of unaligned, roughly cut parts with possibly mismatched joint regions (top row: all parts come from {\em different\/} shapes), 
	the alignment module of our network can learn to arrange these parts into a plausible configuration (middle row). However, this still leaves gaps and geometric/topological mismatches at the joints. 
	After eroding the input parts in advance, 
	our joint synthesis module creates connecting regions, which are stitched to the input to create seamless assemblies (bottom row). 
	}  
	\label{fig:teaser}
	\vspace{-5mm}
\end{figure}

Shape assembly by composing component parts has been the dominant paradigm for data-driven 3D modeling and design exploration~\cite{mitra2013,funkhouser2004,chaudhuri2010,xu2012,ritchie2018}. The component assembly gathers parts from a shape collection and composes them into a target shape. To date, most research effort has been devoted to finding compatible parts for the assembly~\cite{chaudhuri2011,jain2012,kim2013,zheng2013,kalogerakis2012,laga2013,sung2017}. 
However, 
these parts still need to be properly aligned and then connected or joined in a natural and coherent fashion, ideally resulting in a high-quality final shape that is suitable for fabrication, simulation, high-resolution rendering, and seamless texturing.

Compared to the higher-level and human-centric tasks, such as part selection~\cite{chaudhuri2010,sung2017} and assembly exploration~\cite{xu2012}, geometrically aligning and joining 3D parts is a much more tedious and unnatural practice for human modelers. %
%
%
A full automation of this low-level task is highly desirable. However, due to the large variability exhibited by the geometry and topology of the shape parts, 
heuristic approaches \revised{such as hole filling and stitching} are hardly general and reliable, as they must make varying assumptions about part placement, feature alignment and boundary geometry, or resort to user interactions 
\cite{kraevoy2007,duncan2016,huang2012,yu2004,lin2008,sharf2006,schmidt2009}. 
Further, nearly all these analytical methods (except that of Huang et al.~\cite{huang2012}) are primarily designed for organic shapes with smoothness priors, and do not handle man-made shapes with sharp corners, edges, and rich structural variations.

In this paper, we introduce COALESCE, the first {\em data-driven\/} framework for component assembly by learning to {\em align\/} shape parts {\em and synthesize\/} the connections between them. We stipulate that \revised{in the absence of any mathematical formulation of how to connect parts}, the most general and effective way to handle the immensely rich part connection varieties is to {\em learn\/} how parts are connected from a large variety of 3D shapes. In addition, the most viable solution to handle geometric or topological mismatches between two parts is to remove the mismatched portions, via an {\em erosion\/} process, and then rely on a {\em learned joint synthesis\/} step to produce a natural and plausible part connection.

A defining feature of our learning-based part connection framework is that it focuses on {\em joint} synthesis, rather than learning to reconstruct whole shapes. With proper part alignment, which is also learned, the space of part joints possesses a higher level of predictability compared to that of the whole shapes, making joint synthesis a comparably easier learning problem. 
%
%
%
Another novel feature of our method, compared to previous works on hole filling and shape completion, is 
the ``retreat (erode) before advance'' connection mechanism. It can be regarded as a means of normalization to handle the wide variety of possible part connections. 

The main contributions of our work include:
\vspace{-3pt}
\begin{itemize}
  \item We tackle an important problem for assembly-based shape modeling: how to best connect parts, which may come from different 3D models and possess varying degrees of geometric and structural incompatibilities.
  \vspace{-3pt}
  \item An erosion-based part connection mechanism to help reconcile geometric and topological mismatches between parts, and potential part misalignment. 
  \vspace{-3pt}
  \item A novel joint synthesis method to connect input parts into a plausible 3D object, via an implicit surface. As shown in the leftmost chair assembly in Figure~\ref{fig:teaser}, our network is able to connect parts into a novel structure which is absent from the parts' shape origins.
\end{itemize}

Our networks are trained on the dataset of segmented 3D shapes from Yi et al.~\cite{yi2016scalable}, which
was extracted from ShapeNet~\cite{chang2015_shapeNet}. We demonstrate the performance of our ``learning to connect'' framework on a variety of 3D shapes, both qualitatively and quantitatively, through comprehensive comparative studies.

\section{Related work}
\label{sec:related}

\begin{figure*}[!t]
	\centering
	\includegraphics[width=0.9\linewidth]{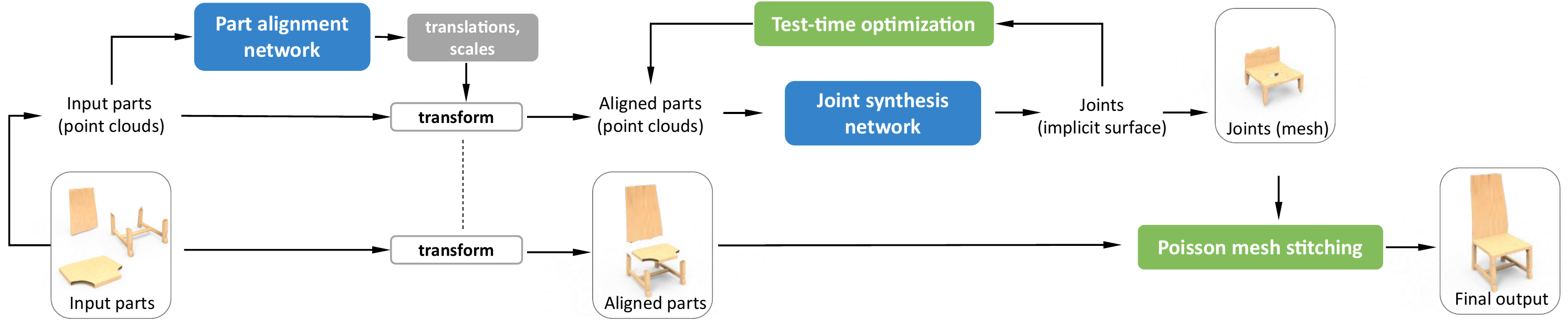}
	\vspace{-2mm}
	\caption{An overview of COALESCE, our component-wise shape assembly. Input mesh parts, with the joint areas eroded to create room for joint synthesis, are first transformed by a part alignment network. Then, a joint synthesis network is employed to synthesize a joint region as an implicit field between each pair of aligned parts. The implicit field is then polygonized and finally Poisson stitching is applied to the input parts and the synthesized joints to produce the final mesh output. Note: dashed lines indicate that the set of part transforms is shared between the point cloud (top) and mesh (bottom) branches.  }
	\label{fig:overview}
	\vspace{-4mm}
\end{figure*}

The literature on modeling shapes from parts is extensive. 
Hence we mainly discuss geometric stitching with analytical approaches, assembly-based modeling from given parts, and modern neural shape synthesis methods.

\vspace{-7pt}
\paragraph{Part stitching.}

Several papers have studied the problem of stitching parts together by connecting a pair of boundary edge loops, where the principal objective is smoothness of the connection. 
Yu et al.~\cite{yu2004} proposed gradient-domain Poisson stitching: after establishing correspondences between the loop vertices, a target continuous gradient field is set up in the adjacent mesh regions. 
Sharf et al.~\cite{sharf2006} proposed a ``Soft-ICP'' algorithm, which iteratively and non-rigidly brings two overlapping patches (one adjacent to each loop) together until they smoothly blend with each other. 
Lin et al.~\cite{lin2008} create a sketch-guided implicit field to connect two boundary loops. The use of an implicit field to generate a connecting surface is similar to our approach, but the synthesis is entirely guided by user input. 
Schmidt and Singh~\cite{schmidt2009} join a boundary edge loop to a target surface by creating consistent local parametrizations on the surface as well as in the polygon bounded by the loop. 
%
\kx{
The works of Huang et al.~\cite{huang2012} and Lescoat et al.~\cite{lescoat2019connectivity} share a more similar goal as COALESCE. 
However, their methods rely on the existence of an underlying smooth transition between the parts that can be detected via geometric analysis. 
}
%
%
\kx{In contrast to these methods, our approach {\em learns} how to repair geometrically and topologically dissimilar 
junctions, generates both low and high curvature regions as indicated by the training data, and co-aligns the assembled parts.}



\vspace{-7pt}
\paragraph{Assembly-based modeling.}
A variety of methods studied ways to construct shapes from pre-existing parts. The pioneering work of Funkhouser et al.~\cite{funkhouser2004} presented tools for searching for database shapes partially matching a user-provided proxy, cutting out a matching part using ``intelligent scissors'', and pasting the part into a new model. The majority of subsequent assembly-based methods focused on retrieving and organizing compatible parts to facilitate or automate assembly \cite{kreavoy2007,chaudhuri2011,jain2012,kalogerakis2012,talton2012,xu2012,fan2016buildings,jaiswal2016,sung2017,ritchie2018,CompoNet,PQNet,Dubrovina2019}. These methods did not place particular emphasis on repairing or synthesizing the connections between parts, especially for man-made shapes. Often, parts were simply placed in contact, or slightly overlapped, to appear visually connected, although Kraevoy et al.~\cite{kreavoy2007}, Chaudhuri et al.~\cite{chaudhuri2011} and Kalogerakis et al.~\cite{kalogerakis2012} did create smooth transitions between organic parts using methods described above.

The SCORES system of Zhu et al.~\cite{zhu2018} shares some aspects of our method: it attempted to realign a crude input arrangement of parts in order to resolve incompatibilities and misalignments
using a recursive neural network. However, this work focused on {\em structural} repair. Only affine transforms were applied to input parts, and no stitching was performed at joints. 
Duncan et al.~\cite{duncan2016} presented a method to deform corresponding joint regions in a shape collection to resemble each other, so that parts could be seamlessly interchanged. However, this required substantial modifications to the input parts, and forced all joints between parts from the same two semantic categories to look identical.
In general, complicated joints, especially in man-made shapes, remain an open problem. 



\vspace{-7pt}
\paragraph{Generative models of 3D geometry.}
In contrast to the assembly-based methods, which {\em reuse} existing shape parts, substantial recent research has focused on {\em synthesizing} the part geometry itself. Neural generative models employ various representations for the output shape, including point clouds \cite{achlioptas18}, voxel grids \cite{wu2015,yumer2016}, octrees \cite{wang2018aocnn}, deformations of a template mesh \cite{Groueix18a}, 3D embeddings of 2D atlases \cite{Groueix18}, or combinations of the above \cite{Muralikrishnan19}. Recent techniques model shapes as continuous implicit functions, where the network learns to map a 3D point to an occupancy value \cite{mescheder2019occupancy} or a clamped signed distance field \cite{Park_2019_CVPR}. We employ a variant of these methods \cite{chen2019learning} in our joint synthesis, as they demonstrate high visual quality and can capture complex topological variations of shapes. Since these implicit functions are defined everywhere in the 3D volume, it is easier for us to incorporate them into consistency losses that ensure the synthesized joint aligns with the input geometry, without expensive closest point sampling. 
\kx{
Some methods explicitly model part layout in addition to generating part geometry \cite{huang2015,li2017,wang2018global,li2019learning,gao2019SDM,SAGnet19,mo2019structurenet,li2020impartass, Huang20GenPartAss,  yang2020dsm}, but none of them explicitly focus on joints between parts.
}

Neural generative models have been commonly used in encoder-decoder architectures; they encode input that is partial or of another modality, such as images or depth scans, and output full 3D shapes. The output is evaluated via reconstructive \cite{fan2017point} or discriminative \cite{3dgan} losses to train the model. Hence, a na\"{i}ve solution to joint synthesis could employ a shape completion approach \cite{dai2017complete}, where the system would encode unstitched parts, and decode the entire shape. Unfortunately, this approach does not preserve the input parts or provide joints of the highest quality, since a lot of the network's capacity is allocated to representing the entire shape, as we show in our experiments. In contrast, our method focuses specifically on synthesizing the joint regions, in a part- and global shape-aware way.

\begin{figure*}[t!]
	\centering
	\includegraphics[width=0.98\linewidth]{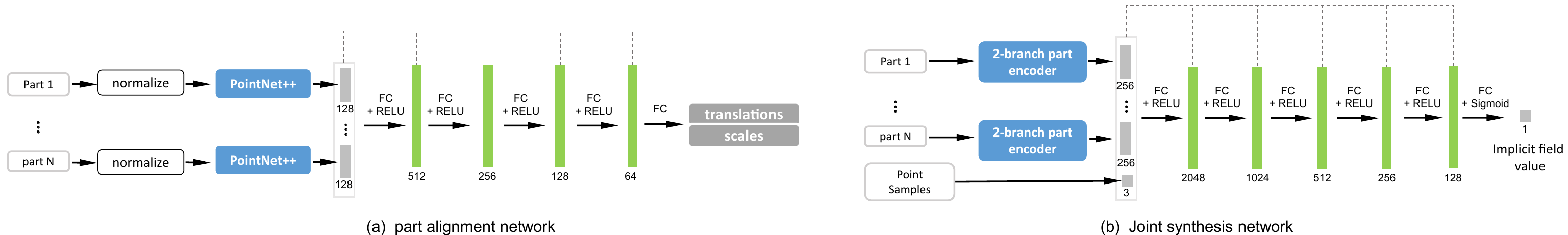}
	\vspace{-2mm}
	\caption{\revised{The architectures of  our part alignment network (a), and our joint synthesis network (b), where the input parts are represented by point clouds. Dashed lines denote skip connections. }}
	\label{fig:archi}
	\vspace{-4mm}
\end{figure*}

\section{Method}
\label{sec:method}


Given a set of input parts extracted from {\em different} objects, our network automatically aligns them and synthesizes plausible joints, connecting the parts into a coherent mesh object. Our pipeline is visualized in Figure~\ref{fig:overview}. 

The input to our pipeline is a set of labeled parts, where each part is represented by a triangle mesh. 
\kx{
We uniformly sample points on the input parts, and erode the joint regions on each part to ensure that there is sufficient room for joint synthesis and to avoid structural mismatching between the crudely segmented parts. Specifically, for each point on an input part, we compute its distance to the segmentation boundary and if the distance is less than a threshold $\tau$, the point will be excluded from the part.
}
%

\zq{
We assume that the input parts are in a canonical orientation, and normalize the parts to the same scale.
}
The alignment network takes normalized parts as inputs and predicts translation and scaling to create structurally sound composition of parts~(Section~\ref{sec:alignment}). 
Next, the joint synthesis network takes the aligned parts as inputs and synthesizes connections between them as an implicit surface (Section~\ref{sec:synthesize}). Note that  implicit representation is especially suitable for joint synthesis, since the network is able to focus on the joint area and allocate its capacity to the geometry near the joints. 
To make the output implicit surface closely conform to the input part boundaries, we jointly refine the predicted spatial transformations for the input parts and the weights of the joint synthesis network to minimize a part-joint matching loss.
After that, the implicit surface produced by the joint synthesis network is discretized into a voxel grid and converted into a mesh using marching cubes~\cite{lorensen1987marching}. 
Finally, we develop a Poisson-based approach, a variant of Poisson blending, to seamlessly stitch the synthesized joint meshes with the original meshes of the aligned input parts. The details of the approach is described in the supplementary material. 

\subsection{Part alignment network}
\label{sec:alignment}

Given a set of input parts from different objects, we sample points for each of them and normalize the resulting point clouds.
After that, we use a part alignment network to assemble them into a shape by applying predicted translation and scaling to each part.
As shown in Figure~\ref{fig:archi}(a), the part alignment network contains PointNet++\cite{qi_nips2017} networks that compute shape features for the input parts. Given the features of the parts, we use a fully connected regression network with skip connections to produce the spatial transformation
that aligns the parts. The features for missing object parts are simply set as zero vectors. 
\zq{ We train the part alignment network with parts belonging to the same objects. }
%
Our loss function is the Earth Mover's Distance between the translated and scaled parts and the ground truth. 
More details of the training process and the architecture of the PointNet++ used in the part alignment network are provided in the supplementary material.  



\subsection{Joint synthesis network}
\label{sec:synthesize}

Our joint synthesis network is a core component of the framework. It takes point clouds of the aligned parts as input and reconstructs the joints as implicit surfaces that connect the parts into a plausible object.  
We show in Figure ~\ref{fig:archi}(b) the architecture of our joint synthesis network.
The network is designed as an encoder-decoder network, where each input part is assigned a two-branch point cloud encoder. 
The first branch of the part encoder is a PointNet++ network for extracting overall shape features from the part. 
The second branch is another PointNet++ network for extracting localized shape features from the points that are close to the joint area. 
We represent each part with 2048 points, and consider the 512 points that are closest to the joint area as the input to the second branch for localized shape features.
The detailed architecture of the two-branch part encoder is provided in the supplementary material.

The network uses an IM-NET~\cite{chen2019learning} to reconstruct implicit surfaces for the joints. Given an object category, for each semantic part segment, we use the two-branch part encoder to extract a feature vector from the part. If a part does not exist in the object, we use a zero vector instead.  The feature vectors of $N$ parts are then concatenated into a single vector to be fed into the IM-NET decoder.

\begin{figure}[t!]
	\centering
	\includegraphics[width=0.75\linewidth]{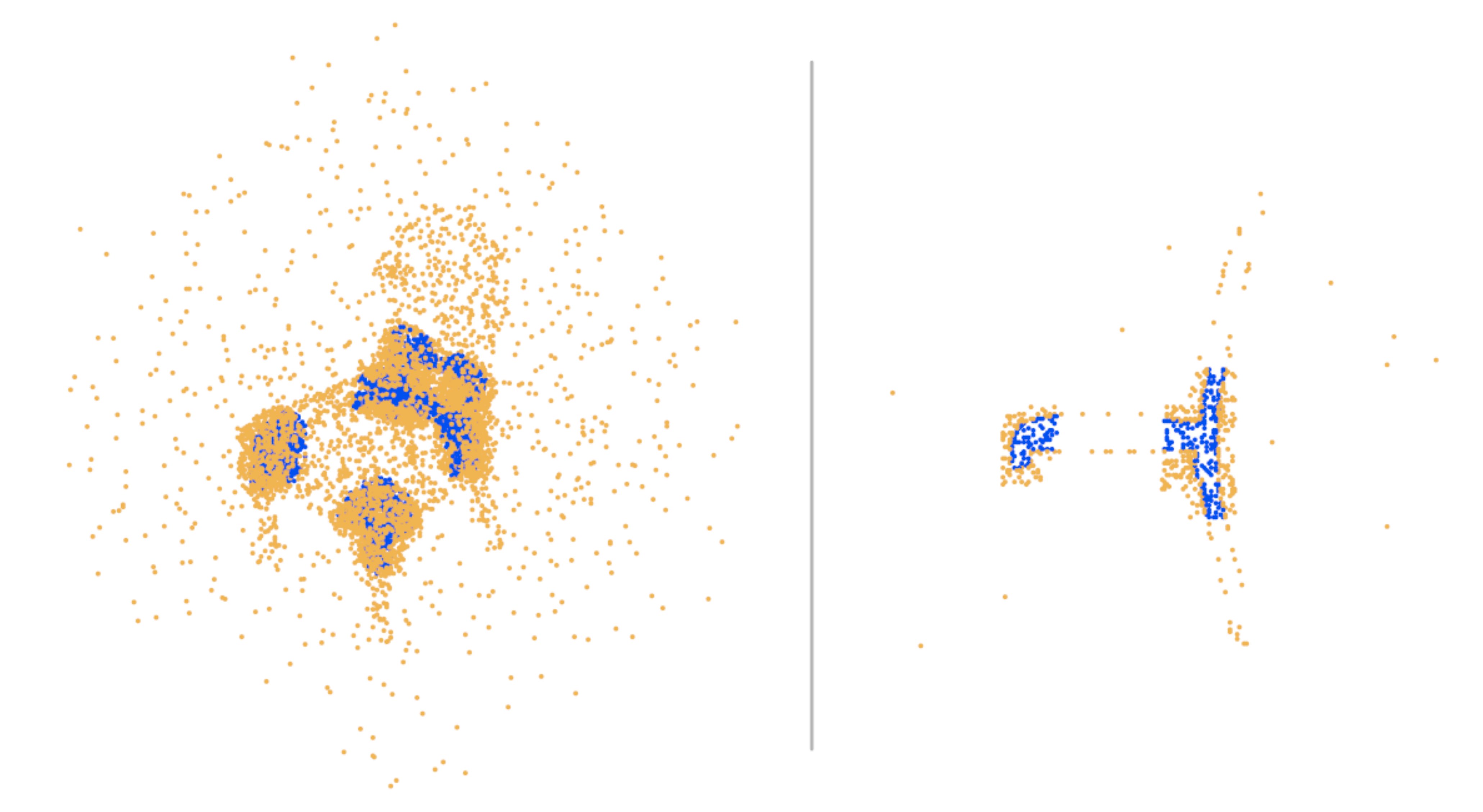}
	\vspace{-3mm}
	\caption{Visualization of our point sampling for training the IM-decoder. Dark yellow dots denote points outside the shape, blue dots denote points inside the shape. 
	}
	\label{fig:sampleVisualize}
	\vspace{-4mm}
\end{figure}

The implicit decoder requires point samples in the 3D space surrounding the shape we aim to reconstruct.  When we sample points with inside/outside status for training the IM-decoder,  to make the network focus on the joint area,  we densely sample points around the joints and sparsely sample points away from the joints as shown in Figure~\ref{fig:sampleVisualize}.  Specifically, we use 16384 sampled points for each shape: 80\% of the points are within the joint volume after 5 steps of volumetric inflation in a $256^3$ voxel grid, and
10\% are outside the shape but close to the surface of the shape, and the remaining 10\% are random samples outside the shape.  By sampling training points in this fashion, we only force the IM-decoder to reconstruct the joint area rather than the entire object.

The loss function of the joint synthesis network is defined by two terms:
\begin{equation*}
\scalebox{0.9}{$L_{\text{joint}} = L_{\text{mse}} + \alpha L_{\text{match}}$}
\end{equation*}
where $\alpha$ is a balancing weight \revised{set to 0.2} in all experiments.
The first term $L_{\text{mse}}$ is the original loss term in  IM-NET which is defined as:
\begin{equation*}
\scalebox{0.9}{$L_{\text{mse}} = \frac{1}{|S|} \sum_{p\in S}{ | f(p) - F(p) |^2} $}
\end{equation*}
where $S$ is the set of point samples, $f(p)$ is the output value from our network for point sample $p\in S$. $F(p)$ is the ground-truth label of point $p$.

The term $L_{\text{match}}$ is to encourage the synthesized joint to match the open boundaries of the input parts.  To define $L_{\text{match}}$,  we first select a set of 1024 points $\mathcal{N}(J)$ that are on the surface of the input parts and close to the joint. If the implicit surface tightly matches the open boundaries of the input parts, for each point $p \in \mathcal{N}(J)$, if we move it along its normal direction, the expected implicit function value on its new position is 0 (outside). If we move it in the inverse direction of its normal direction,  the expected value is 1 (inside).  Thus we define $L_{\text{match}}$ as: 
\begin{equation*}
\scalebox{0.9}{$L_{\text{match}} = \frac{1}{2|\mathcal{N}(J)|} \sum_{p\in\mathcal{N}(J) }{|f(p + \lambda n)|^2  + |f(p-\lambda n)-1|^2}$}
\end{equation*}
where $n$ is the unit normal vector of $p$, and $\lambda$ is a small constant with a default value of 0.005.

The joint synthesis network is trained by segmented parts of 3D objects. To make room for the network to learn to synthesize the joints connecting input parts, we erode the parts along the segmentation boundaries. As discussed in section~\ref{sec:intro}, another effect of the erosion is to help eliminating topological mismatches among input parts from different objects.
%
We first pre-train the two-branch point cloud encoders for 100 epochs by pairing it with a fully connected decoder~\cite{achlioptas18} for reconstructing 3D shape parts.
Afterwards, we pair the pretrained point cloud  encoders with the IM-decoder, and train the IM-decoder with the loss function $L_{\text{joint}} = L_{\text{mse}}$ for 80 epochs. 
Finally, we train the encoders and the IM-decoder together with the loss function $L_{\text{joint}} = L_{\text{mse}} + \alpha L_{\text{match}}$ for another 80 epochs. 
More details on our training procedures and training time can be found in the supplementary material.


\subsection{Test-time optimization}
\label{sec:optimize}

It is critical that the synthesized joints align closely with the open boundaries of the parts to create a realistic model. While our feed-forward joint synthesis network attempts to maximize boundary consistency, to further improve the results, we employ test-time optimization to explicitly minimize any boundary discontinuities. In this phase, we jointly optimize the scaling factors $\{s\}$, translations $\{t\}$ of parts, and the parameters $W$  of the IM-decoder in the joint synthesis network for a given test input.
We define the objective function for this optimization as an $L_1$ version of the part-joint matching term:
\begin{equation*}
\scalebox{0.85}{$h(\{s\}, \{t\}, W ) = \frac{1}{2|\mathcal{N}(J)|} \sum_{p\in\mathcal{N}(J) }{|f(p + \lambda n)|  + |f(p-\lambda n)-1|}$}
\end{equation*}
where ${N}(J)$ is a set of 1024 points close to the joint, $n$ is the unit normal at $p$, and $\lambda$ is a small constant set to 0.005.

Minimizing the objective function $h$ with respect to $\{s\}$, $\{t\}$ requires a significantly different step size compared to minimizing the objective function with respect to network parameters $W$. This motivated our choice to alternately optimize for part alignments and joint synthesis.  
\revised{
We first fix the parameters of the IM-decoder and optimize  $\{s\}$, $\{t\}$ by minimizing the objective function $h$ with 
an Adam optimizer with step size 0.002.
After that, we fix  $\{s\}$, $\{t\}$ and optimize  the parameters of the IM-decoder  $W$ with step size 0.0001. 
We keep repeating the above two steps for a fixed number of iterations (25, in our implementation).
}
As we show in Section~\ref{sec:ablation}, the test-time optimization refines the positions and scales of the parts and the decoding result such that they match more tightly.

\begin{figure*}[t!]
	\centering
	\includegraphics[width=0.9\linewidth]{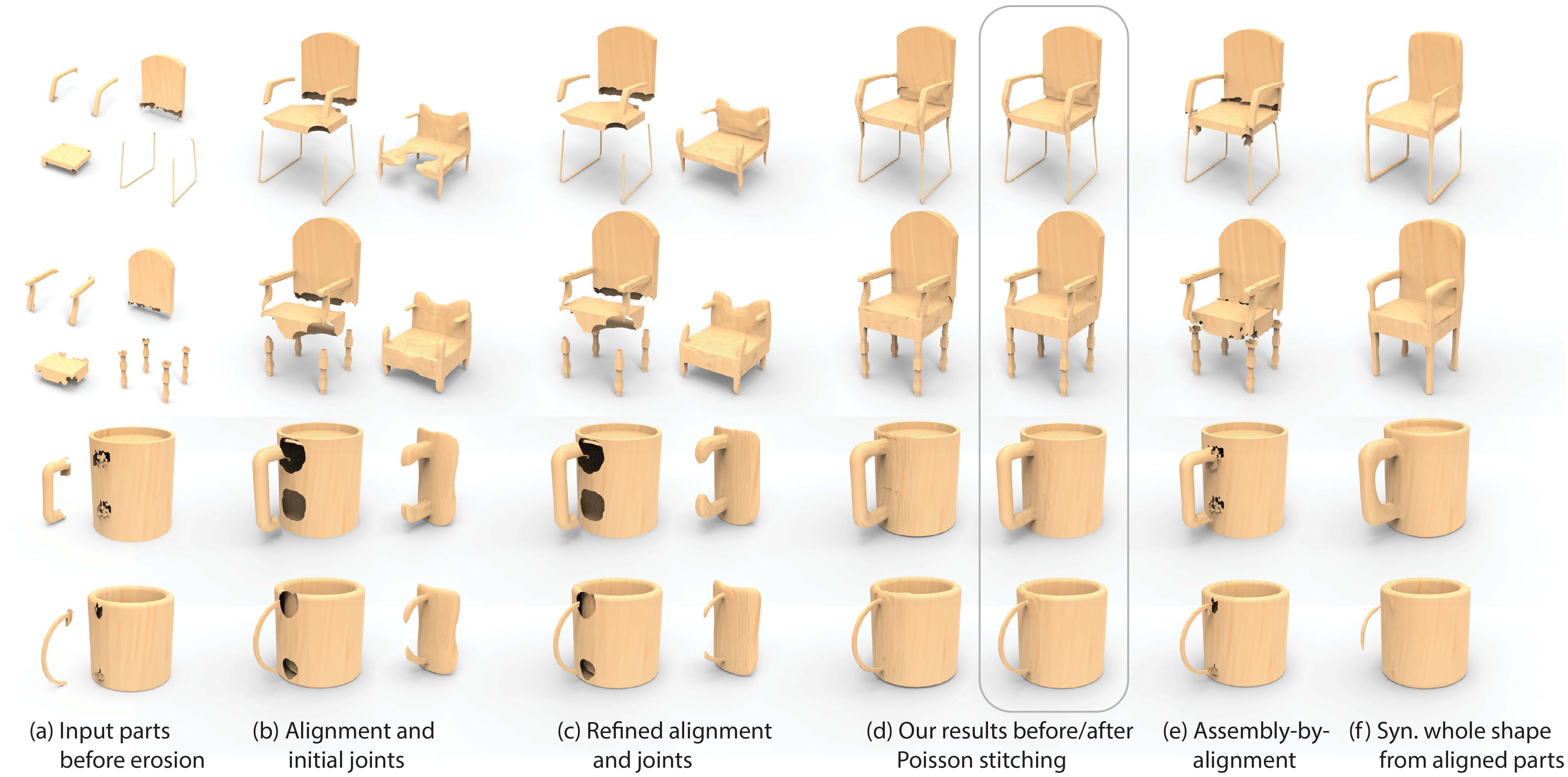}
	\vspace{-2mm}
	\caption{\revised{We evaluate the significance of different components of our framework by ablation. Given the input parts (a), we show results that only involve part alignments (b), refined alignment with test-time optimization (c), and the effect of Poisson stitching (d). We also demonstrate the disadvantages of naive approaches, such as assembling input parts without any joint synthesis (e), or directly synthesizing the entire shape  (f) from aligned parts. } }
	\label{fig:ablation}
	\vspace{-3mm}
\end{figure*}

\subsection{Poisson mesh stitching}

After test-time optimization, the joint region represented by implicit function is converted into a polygon mesh by Marching Cubes. A Poisson-based approach is developed to seamlessly stitch the synthesized mesh with the aligned parts. The Poisson mesh stitching has three phases: (i) finding corresponding loops between the input parts and the synthesized joints,  (ii) removing redundant regions from the joint mesh and (iii) Poisson mesh blending. Please find the description of these steps in the supplementary material.
\section{Experiment and results}
\label{sec:results}

We evaluate the utility of COALESCE for the part assembly problem. In the following, we discuss our data pipeline, ablation studies for various design choices, and comparisons to alternative approaches. 
More visual results can be found in the supplementary material.
Unless stated otherwise (e.g., in ablations), we use the same network architecture with the same hyper-parameters and the same Poisson mesh stitching post-processing for all experiments.

\subsection{Dataset}
\label{sec:dataprocessing}

We use three categories of objects from the segmented shapes provided by Yi et al.~\cite{yi2016scalable}: 3,746 chairs, 2,690 airplanes, and 184 mugs. We split each category into training (80\%) and test (20\%) sets. 
We then create eroded part-wise point clouds as described in Section~\ref{sec:method}. For our erosion process we set the erosion radius to $\tau=0.05$ of the bounding box diameter for chairs and mugs and $\tau=0.025$ for airplanes.            
Since there are very few mugs, we  train the joint synthesis network for the mug dataset for 80 more epochs.

\subsection{Ablation study}
\label{sec:ablation}


\paragraph{Qualitative study.}
As shown in Figure~\ref{fig:ablation}, the inputs to our part alignment network are crudely segmented parts from \textit{different} objects. 
Figure~\ref{fig:ablation}b shows that the results of our network may be inaccurate. Similarly, the geometry created by the forward pass through the joint synthesis network might not align tightly with the input parts, e.g., the last row. 
Our test-time optimization addresses these issues and further refines the alignment and the joint geometry (Figure~\ref{fig:ablation}c) to produce the final result (Figure~\ref{fig:ablation}d). 

A naive approach to assemble parts from different objects is to learn to align them directly, without erosion or joint synthesis.  
We train a part alignment network that directly aligns input parts to get the assembly, and show the results in Figure~\ref{fig:ablation}e. We see several artifacts emerging from this naive pipeline. First, there could be a topological or geometric mismatch, such as armrests of a chair falling outside the range of the seat (see second row) or the cup of a mug having holes that are larger than the size of the handle (see last row). Second, segmented parts might have imperfect boundaries (see first row). Note how our erosion and synthesis approach solves these issues (Figure~\ref{fig:ablation}d).

Another alternative is to treat it as a shape completion problem, which is commonly addressed with encoder-decoder architectures. We train a neural network that encodes point clouds of aligned parts, and then decodes them into a full shape via IM-decoder (with the identical architecture to our joint synthesis network). As we can see in Figure~\ref{fig:ablation}f, the results of this approach might not preserve the geometry of the input parts and often have more artifacts near the joints due to the absence of joint focus.

\begin{table}[t!]
\begin{center}
\resizebox{\linewidth}{!}{%
\begin{tabular}{l |r|r|r |r|r|r}
\hline
  & \multicolumn{3}{|c|}{Chamfer Dis. ($\times 10^3$)} & \multicolumn{3}{c}{Light Field Dis.} \\
\hline
  & plane & chair & mug & plane & chair & mug \\
\hline

Synthesize entire shape  & 0.164 & 0.738 & 0.804     & 3265.2 &   3054.3 & 1117.8 \\
Syn. entire shape + code opt.  & 0.173 & 0.472 & 0.646     & 2991.0 &   2374.8 & 885.1 \\
Ours (before test-time opt.)             & 0.073 & 0.225 & 0.297     & 1148.3 & 1251.6 & 374.3  \\
Ours (after test-time opt.)              & 0.071 & 0.191 & 0.300     & 1072.9 & 951.5 & 276.9  \\
Ours (after Poisson blending)   & {\bf 0.067} & {\bf 0.181} & {\bf 0.274} & {\bf 895.4} &  {\bf 894.6} & {\bf 196.2} \\

\hline
\end{tabular}
}
\end{center}
\caption{
\kx{Quantitative ablation study. We compute the errors of our method against ground-truth shapes on different pipeline stages including before and after test-time optimization, and after Poisson blending.  
}
}
\vspace{-4mm}
\label{table:quan_ablation}
\end{table}

\vspace{-7pt}
\paragraph{Quantitative study.}
In addition to the qualitative study in Figure~\ref{fig:ablation}, we also did a quantitative ablation study in Table~\ref{table:quan_ablation}.
In the study, we feed the eroded input parts from the same test shape to our network, and measure the chamfer distance ({\em lower} is better) and light field distance (LFD: {\em lower} is better) between the results and the ground-truth test shapes. The quantitative evaluation results show
that the test-time optimization module and the Poisson blending module can improve the quality of the network output.
\zq{
We also evaluate the results of a network that directly synthesizes the entire shape from the aligned parts (first row in Table~\ref{table:quan_ablation}), and optimizes the latent code to minimize a matching loss that is similar to $h(\{s\}, \{t\}, W)$ but on the entire shape rather than just joints (second row in Table~\ref{table:quan_ablation}).
}

\begin{figure}[t!]
	\centering
	\includegraphics[width=1.0\linewidth]{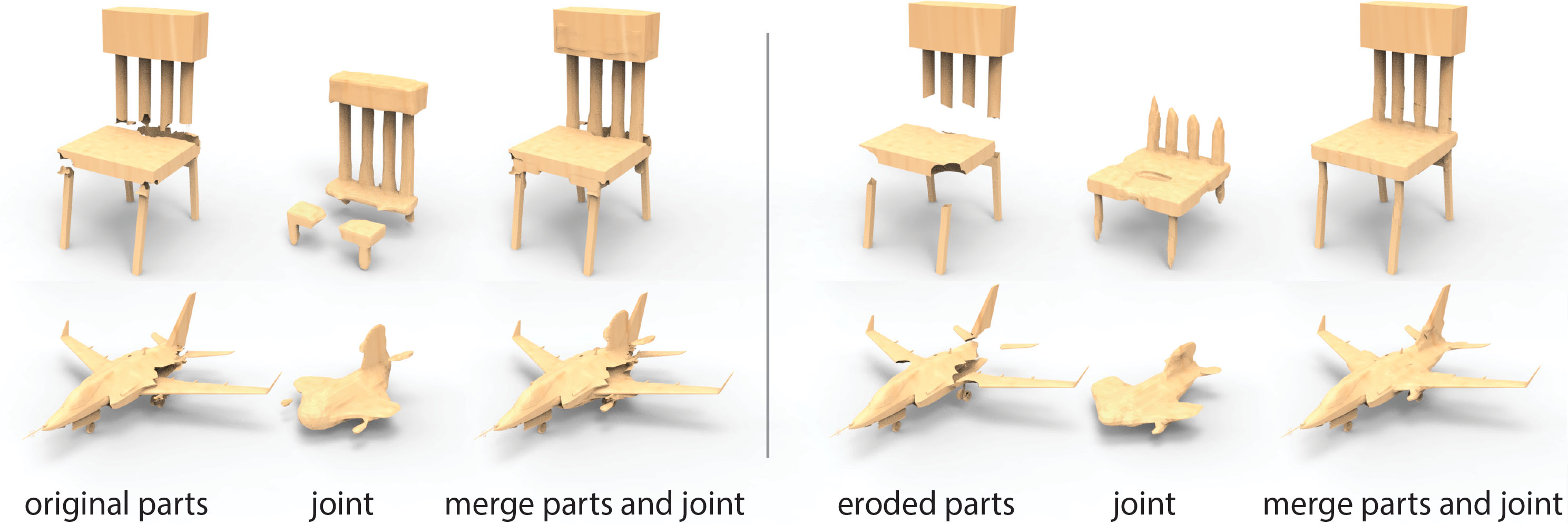}
	\vspace{-2mm}
	\caption{\revised{Comparison of two networks that synthesize part connections from original parts before erosion (left) and from eroded parts (right). Incorporating erosion better resolves geometric and topological mismatches.} }
	\label{fig:erosion}
	\vspace{-4mm}
\end{figure}

\vspace{-7pt}
\paragraph{Part erosion.}
To show the necessity of part erosion, we train a neural network to synthesize joints from {\em uneroded} parts, with the same architecture and configuration as our network described in Section~\ref{sec:method}.
We compare the results of directly synthesizing joints from uneroded parts with our results in Figure~\ref{fig:erosion}. Note that the variant without erosion cannot successfully resolve the mismatches at part junctions.

\subsection{Comparison to Baselines}
\label{sec:comparison}
\begin{figure*}[t!]
	\centering
	\includegraphics[width=0.9\linewidth]{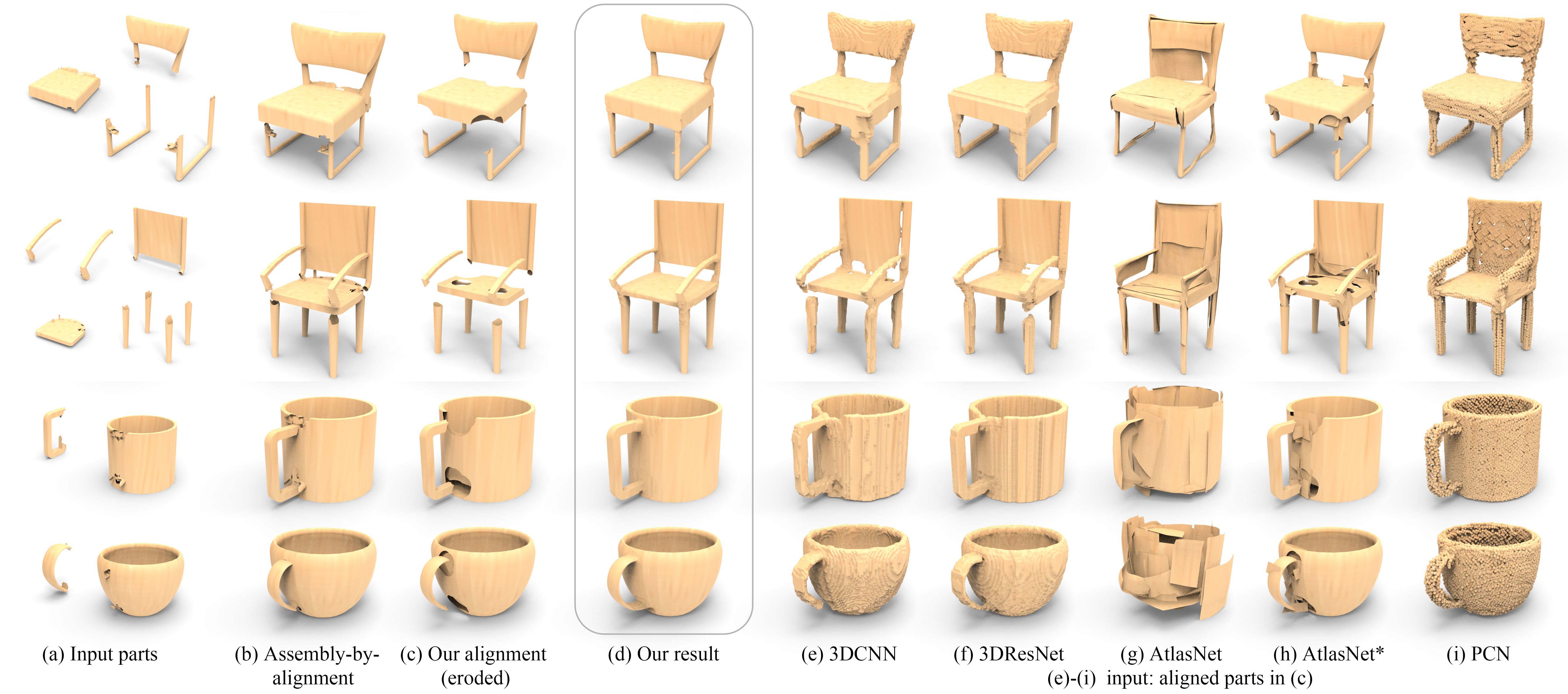}
	\vspace{-2mm}
	\caption{\revised{Qualitative comparisons with baseline methods, which took the parts aligned by our method (c) as input.}}
	\label{fig:compare}
	\vspace{-4mm}
\end{figure*}

Although there are no prior works on learning joint synthesis, we can compare to shape completion methods, such as AtlasNet~\cite{Groueix18} and PCN~\cite{yuan2018pcn}. We directly use the publicly available source codes of these methods and train the networks on individual categories following the provided instructions. We also tested a customized version of AtlasNet that only outputs surfaces in the joint regions, similar to our joint synthesis network. We denote this version of AtlasNet as AtlasNet* in Figure~\ref{fig:compare}h and Tables~\ref{table:numbers_gt_aligned},~\ref{table:numbers_random_aligned},~\ref{table:numbers_sim_aligned}.

We also consider two voxel-based shape completion baselines that can be applied to our problem: 3DCNN and 3DResNet. In the two baselines, we use architectures that perform shape completion akin to image translation tasks. See the supplementary for more details. 
Unlike encoder-decoder architectures such as AtlasNet and PCN, these methods do not have a tight bottleneck layer, and they only modify inputs in a relatively local manner. Thus, they do not have the challenge of reconstructing the entire shape from a compact latent code, and might be better at preserving the geometry of the original parts. When testing, we use marching cubes to obtain the meshes from the output voxels.

The qualitative comparison of our method to the baselines is provided in Figure~\ref{fig:compare}. The input parts in Figure~\ref{fig:compare} are from different objects (Figure~\ref{fig:compare}a), and naively aligning them (Figure~\ref{fig:compare}b) results in a mesh with a large number of boundary artifacts. We feed eroded parts (Figure~\ref{fig:compare}c) as input to our method and various baselines. Our method produces high quality seamless meshes (Figure~\ref{fig:compare}d). Voxel translation techniques (3DCNN, 3DResNet) suffer from discretization artifacts and often produce poor joints (Figure~\ref{fig:compare}e,f). Encoder-decoder architectures (AtlasNet, PCN) tend to significantly change the input geometry (Figure~\ref{fig:compare}g,i). While modified AtlasNet* preserves the original geometry by construction, the synthesized joints do not seamlessly merge with the rest of the geometry (Figure~\ref{fig:compare}h).

\begin{table}[t!]
\begin{center}
\resizebox{0.9\linewidth}{!}{%
\begin{tabular}{l |r|r|r |r|r|r}
\hline
  & \multicolumn{3}{|c|}{Chamfer Distance ($\times 10^3$)} & \multicolumn{3}{c}{Light Field Distance} \\
\hline
  & plane & chair & mug & plane & chair & mug \\
\hline

AtlasNet   & 0.117 & 0.402 & 0.885 & 2402.1 & 2482.4 & 2227.9  \\
AtlasNet*   & {\bf 0.067} & 0.205 & 0.277 & 1049.4 & 1316.8 & 455.2  \\
PCN        & 0.095 & 0.308 & 0.506 & - & - & -  \\
3DCNN      & 0.151 & 0.268 & 0.440 & 2464.6 & 1446.2 & 615.8  \\
3DResNet   & 0.141 & 0.262 & 0.421 & 2086.2 & 1313.2 & 426.9  \\
Ours       & {\bf 0.067} & {\bf 0.181} & {\bf 0.274} & {\bf 895.4} & {\bf 894.6} & {\bf 196.2}  \\

\hline
\end{tabular}
}
\end{center}
\caption{
Quantitative results of different methods when the input parts are from the same object and aligned with respect to the ground truth positions. Chamfer Distance is multiplied by $10^3$. Light Field Distance cannot be applied to PCN since PCN generates point clouds rather than meshes.
}
\vspace{-3mm}

\label{table:numbers_gt_aligned}
\end{table}
\begin{table}[t!]
\begin{center}
\resizebox{0.9\linewidth}{!}{%
\begin{tabular}{l |r|r|r |r|r|r}
\hline
  & \multicolumn{3}{|c|}{Chamfer Distance ($\times 10^3$)} & \multicolumn{3}{c}{Light Field Distance} \\
\hline
  & plane & chair & mug & plane & chair & mug \\
\hline

AtlasNet   & 0.200 & 0.525 & 0.948 & 3462.5 & 2691.9 & 2240.4  \\
AtlasNet*   & 0.092 & 0.258 & 0.288 & 1716.5 & 1517.9 & 485.5  \\
PCN        & 0.173 & 0.489 & 0.583 & - & - & -  \\
3DCNN      & 0.199 & 0.330 & 0.495 & 2896.6 & 1563.2 & 655.4  \\
3DResNet   & 0.179 & 0.319 & 0.485 & 2265.7 & 1414.0 & 449.2  \\
Ours       & {\bf 0.080} & {\bf 0.222} & {\bf 0.287} & {\bf 1302.2} & {\bf 1039.9} & {\bf 198.7}  \\

\hline
\end{tabular}
}
\end{center}
\caption{
\revised{Quantitative results of different methods when the input parts are warped by a sine function.}
}
\vspace{-3mm}

\label{table:numbers_random_aligned}
\end{table}
\begin{table}[t!]
\begin{center}
\resizebox{0.9\linewidth}{!}{%
\begin{tabular}{l |r|r|r |r|r|r}
\hline
  & \multicolumn{3}{|c|}{Chamfer Distance ($\times 10^3$)} & \multicolumn{3}{c}{Light Field Distance} \\
\hline
  & plane & chair & mug & plane & chair & mug \\
\hline

AtlasNet   & 0.313 & 0.666 & 1.394 & 4158.8 & 3035.8 & 2843.1  \\
AtlasNet*  & 0.109 & 0.255 & 0.302 & 2264.6 & 1835.1 & 737.2  \\
PCN        & 0.274 & 0.512 & 0.989 & - & - & -  \\
3DCNN      & 0.219 & 0.314 & 0.486 & 3682.1 & 1570.3 & 719.8  \\
3DResNet   & 0.177 & 0.304 & 0.476 & 2362.7 & 1403.7 & 505.9  \\
Ours       & {\bf 0.077} & {\bf 0.190} & {\bf 0.292} & {\bf 1247.9} & {\bf 1063.0} & {\bf 231.9}  \\

\hline
\end{tabular}
}
\end{center}
\caption{
\revised{Quantitative results of different methods when the input parts are randomly scaled and translated.}
}
\vspace{-3mm}

\label{table:numbers_sim_aligned}
\end{table}

To quantitatively evaluate our joint synthesis method, we take eroded input parts from the same test shape, and use that test shape as the ground truth.
We report chamfer distance and light field distance as error metrics in Table~\ref{table:numbers_gt_aligned}. Our results have the lowest errors in both measures. Although the customized AtlasNet* has errors close to ours in terms of chamfer distances, it does not have the same visual quality as our results, which is more evident from LFD.

To evaluate the robustness of our method and the baselines to part misalignment, we warp the test input parts and the ground-truth shape globally along the y-axis with a sine transform:
$y \leftarrow y + 0.02 \sin( 4 \pi z + f )$, where $y$ and $z$ are the coordinates of a 3D point $(x, y, z)$,  $f$ is a per-shape random number sampled uniformly from $[-\pi, \pi]$, and $0.02$ is the constant amplitude. 
\kx{
The sine transform creates non-rigid shearing on the input parts, and moves open boundaries of the parts up/down with different amount of translations such that some misalignment is created.}
We report the quantitative performance of our method and the baselines in Table~\ref{table:numbers_random_aligned}.

To evaluate the robustness of our method and the baselines to translation and scaling of the test shapes, \zq{ we apply a global random scaling and translation to the input parts}, with a uniformly-sampled scale in the range $[0.9, 1.1]$, and uniformly-sampled displacements in the range $[-0.04, 0.04]^3$. Table~\ref{table:numbers_sim_aligned} shows the results under such random scaling and translation, demonstrating that our method is the most robust.

\subsection{Generalizability}

\begin{figure}[t!]
	\centering
	\includegraphics[width=0.8\linewidth]{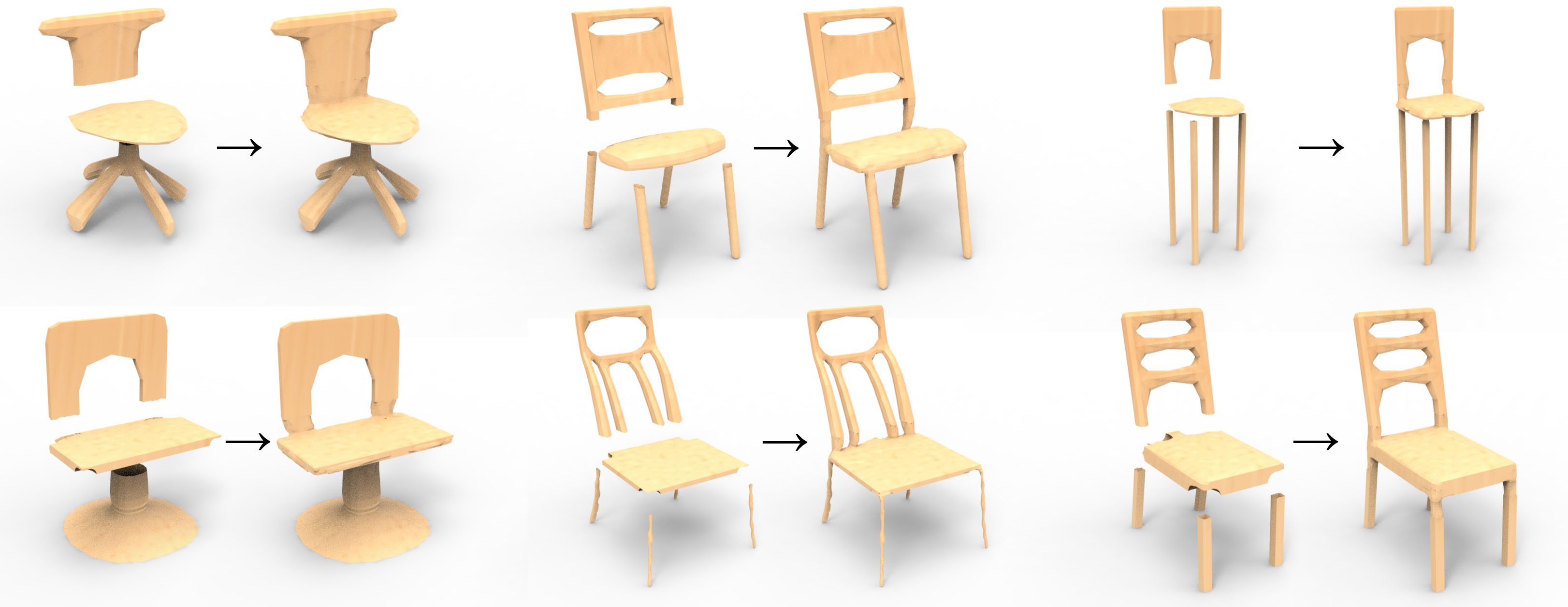}
	\caption{Qualitative results of our method trained on chairs from ShapeNet but tested on chairs from COSEG.}
	\label{fig:cochair}
	\vspace{-2mm}
\end{figure}

We also tested our network on the chair category of the COSEG~\cite{wang2012active} dataset. We do not re-train our network to fit the dataset. Instead, we use the network trained on the chair category from ShapeNet, and test it on COSEG to demonstrate how well it generalizes to novel datasets. Figure~\ref{fig:cochair} shows some results to illustrate such generalizability.

\section{Discussion, limitation, and Future Work}

\begin{figure}[t!]
	\centering
	\includegraphics[width=0.8\linewidth]{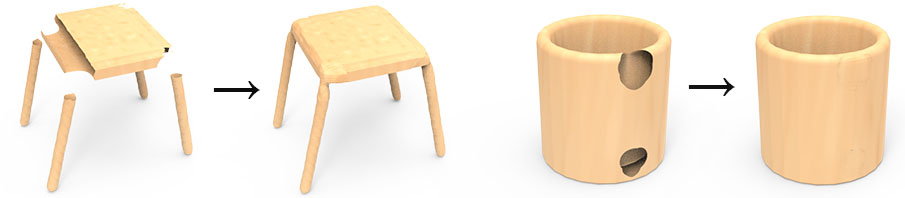}
	\caption{\kx{Our method can produce reasonable results even when a few parts, e.g., a chair back, are absent.}}
	\label{fig:joint}
	\vspace{-4mm}
\end{figure}

We presented COALESCE, the first neural network-based pipeline for aligning and seamlessly connecting parts in a 3D shape assembly. 
In interactive modeling scenarios, it is hard to perfectly align parts, and even harder (and often impossible) to make topologically and geometrically compatible cuts in adjacent parts for the purpose of seamless joining. COALESCE erodes crudely cut joint regions to prepare them for connections, corrects alignment errors in the assembly, and generates seamless connecting patches filling in the eroded joints, accounting for geometric and topological mismatches. Experiments show that our approach significantly outperforms alternatives, and is a promising addition to the assembly-based modeling toolbox, applying modern neural synthesis techniques to address a need largely neglected by prior work.
\kx{Our method is also potentially useful for other applications, such as part-based 3D scanning of objects. 
}

\begin{figure}[t!]
	\centering
	\includegraphics[width=0.9\linewidth]{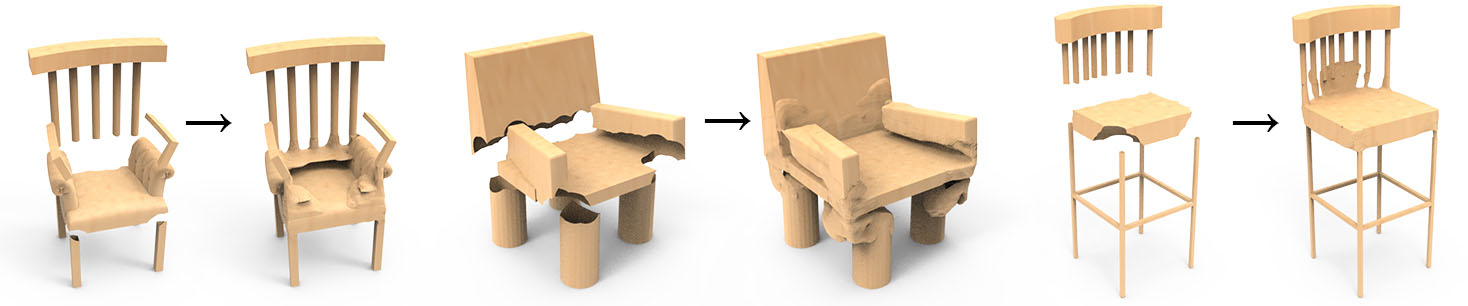}
	\caption{\kx{Examples of failure cases by our approach.} }
	\label{fig:failure}
	\vspace{-4mm}
\end{figure}

\kx{
By simply setting the latent codes of absent parts as zero vectors, our network can be trained for objects with different part configurations. 
In the first two rows of Figure~\ref{fig:compare}, we show that the same network trained with ShapeNet chairs can synthesize joints for both armed and armless chairs. 
The same network can also predict joints for a stool, or fill the holes on a mug when the handle is not provided, as shown in Figure~\ref{fig:joint}. 
However, as one may expect, our network can fail to produce a natural part connection when the geometric and/or topological incompatibility between the parts becomes too significant or goes beyond what may be learned from data. Figure~\ref{fig:failure}-left shows one such case, where erosion cannot sufficiently alleviate the part mismatch. 
Significant part incompatibility can also cause inaccurate alignment and lead to failure; see Figure~\ref{fig:failure}-middle. 
Further, our method is limited by the capability of the PointNet++ encoders and the IM-decoder. When the input parts have complex thin structures, our method may not be able to synthesize detailed structures, e.g. Figure~\ref{fig:failure}-right. 
In addition, the current version of our network may not generalize well to handle fine-level parts. One reason is that our architecture requires the latent codes from different parts to be concatenated. As the part count increases, our network will require more neurons, weights and GPU memory.  Furthermore, fine-level part structures are much less consistent across different objects. Our network may not have sufficient capacity to adapt.
}

Our method is only a first step and still has several limitations, which suggest ripe avenues for future work. First, like most learning-based methods for shape analysis, COALESCE is still category-specific, and relies on predictable, labeled part layouts and joint locations. 
Also, our method has been trained on parts which are in a canonical orientation and compatibly scaled. Furthermore, we use heuristic erosion of the input parts, but we believe it is possible to extend this approach to learn a natural erosion operation. 
%
%
%

\paragraph{Acknowledgements}
This research was supported in part by grants from NSERC and gifts from Adobe Research. We thank members of SFU GrUVi lab and Adobe Research for helpful discussions. Part of this work was done while Kangxue was an intern at Adobe Research.

{\small
\bibliographystyle{ieee}
\bibliography{shape_modeling}
}

\newpage

\appendix
\section{Supplementary material}
\label{sec:appendix}

\subsection{Qualitative results}

\begin{figure*}[h!]
	\centering
	\includegraphics[width=0.9\linewidth]{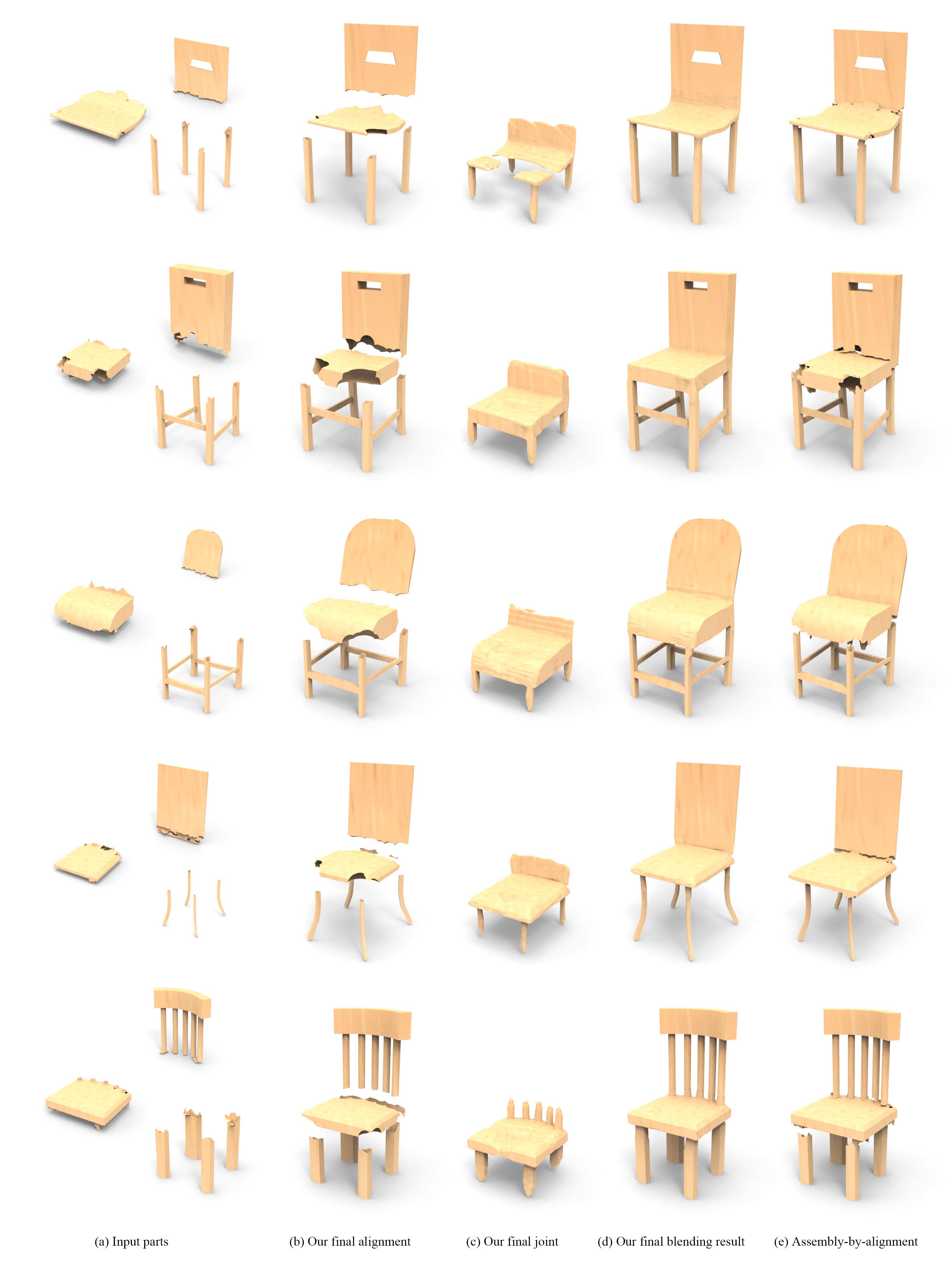}
	\caption{ More examples of chairs.}
	\label{fig:chair3}
\end{figure*}

\begin{figure*}[h!]
	\centering
	\includegraphics[width=0.9\linewidth]{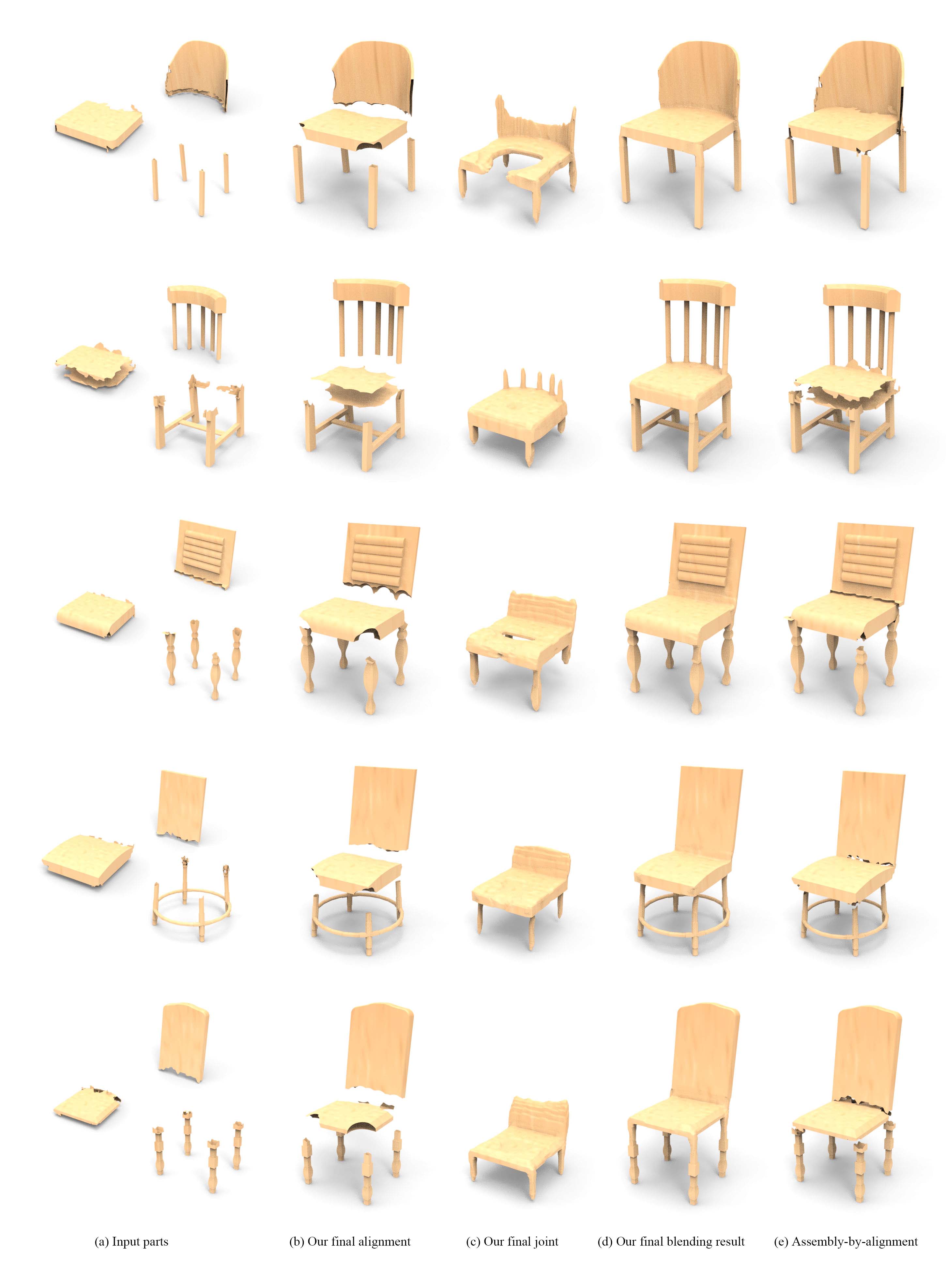}
	\caption{ More examples of chairs.}
	\label{fig:chair4}
\end{figure*}

\begin{figure*}[h!]
	\centering
	\includegraphics[width=0.9\linewidth]{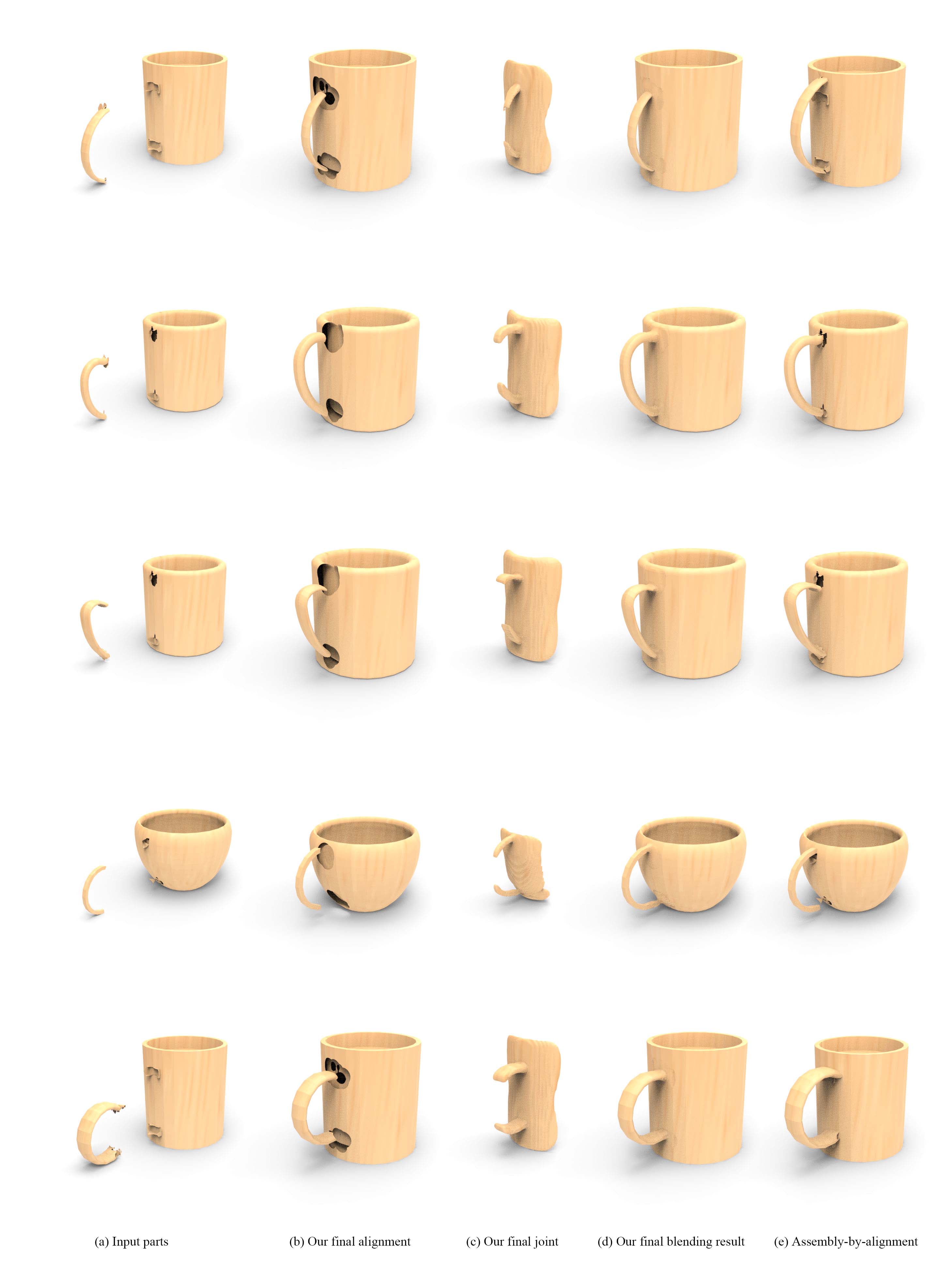}
	\caption{ More examples of mugs.}
	\label{fig:mug2}
\end{figure*}

\begin{figure*}[h!]
	\centering
	\includegraphics[width=0.9\linewidth]{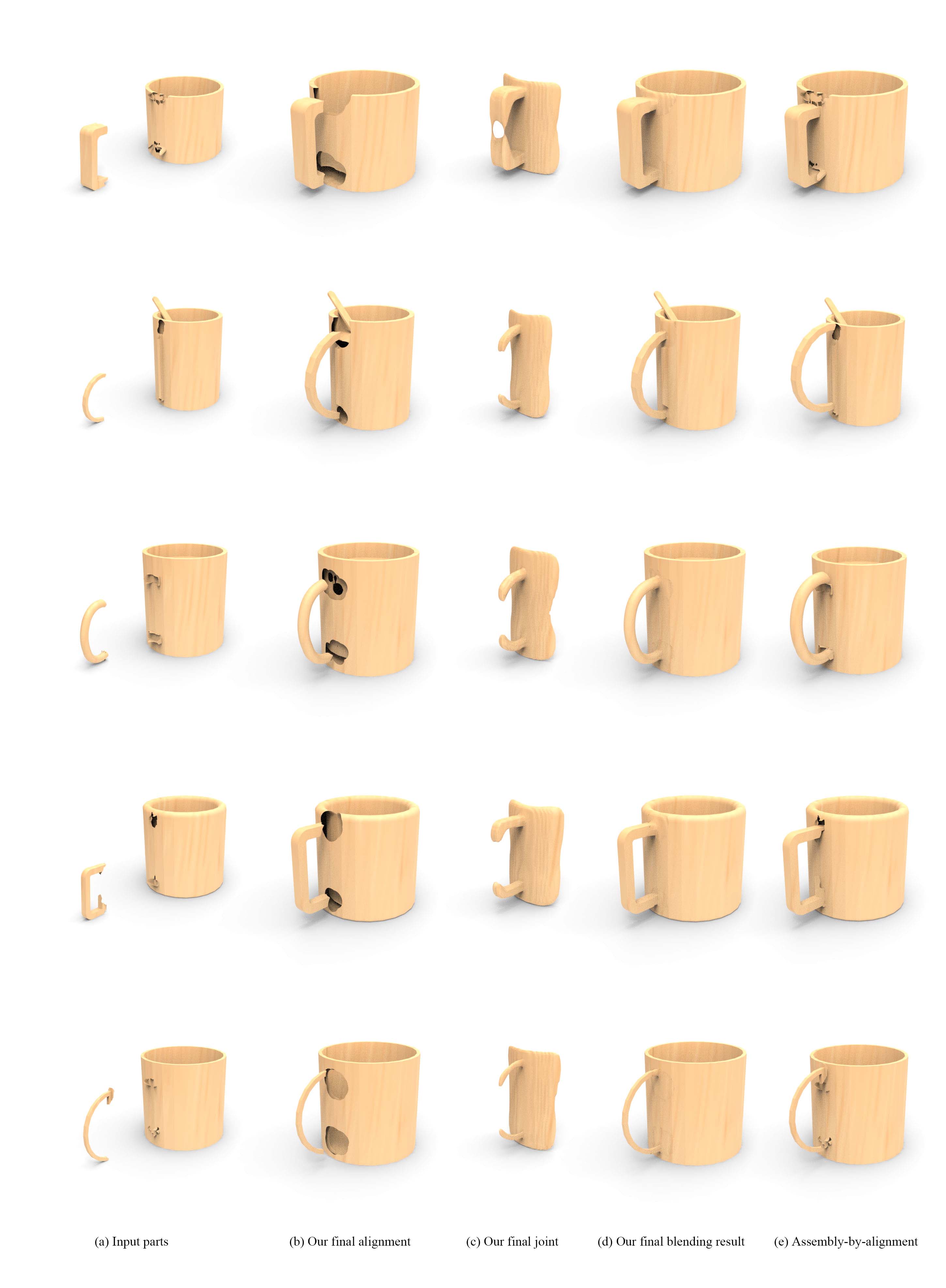}
	\caption{ More examples of mugs.}
	\label{fig:mug4}
\end{figure*}

\begin{figure*}[h!]
	\centering
	\includegraphics[width=0.9\linewidth]{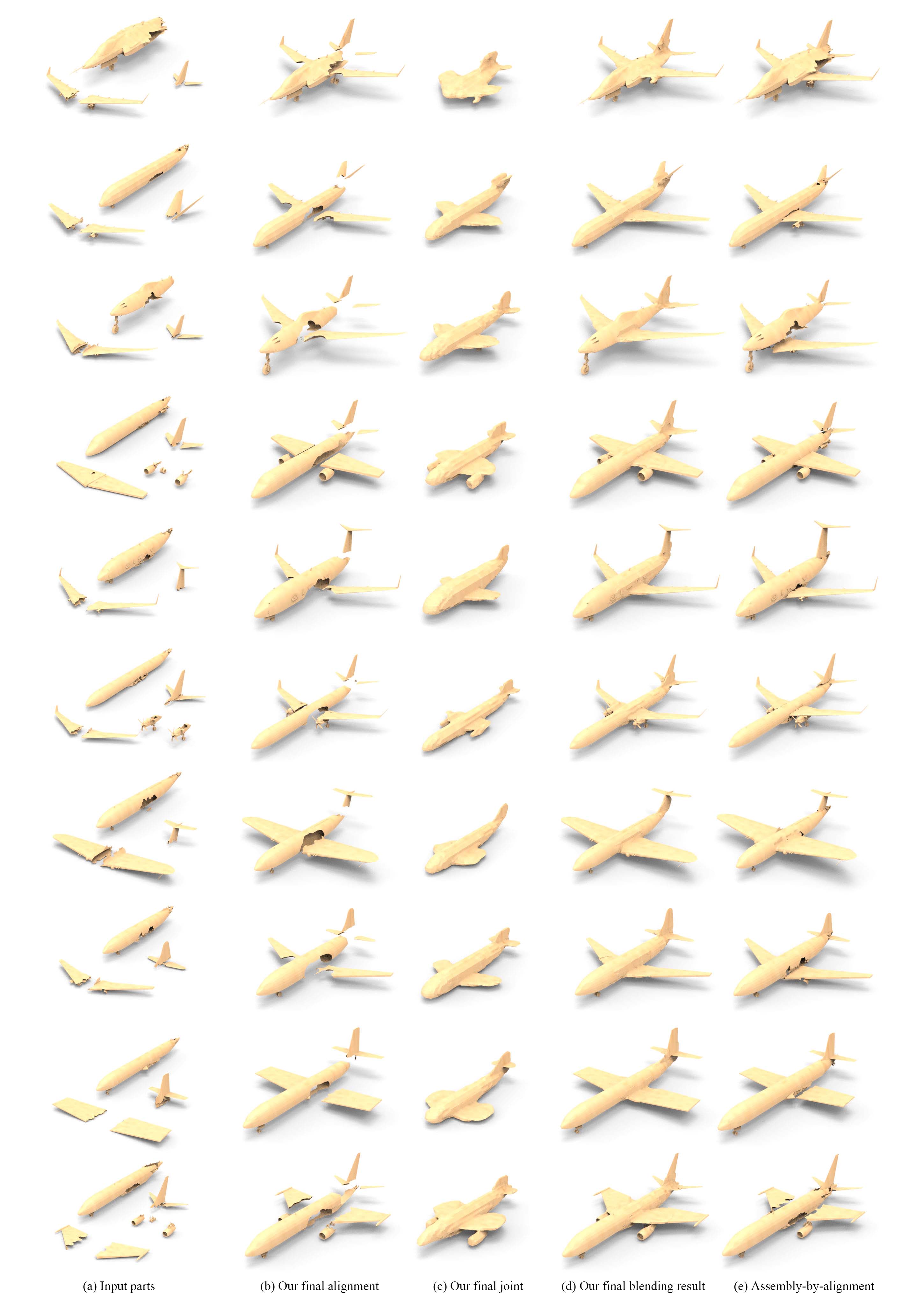}
	\caption{ More examples of airplanes.}
	\label{fig:airplane1}
\end{figure*}

\begin{figure*}[h!]
	\centering
	\includegraphics[width=0.9\linewidth]{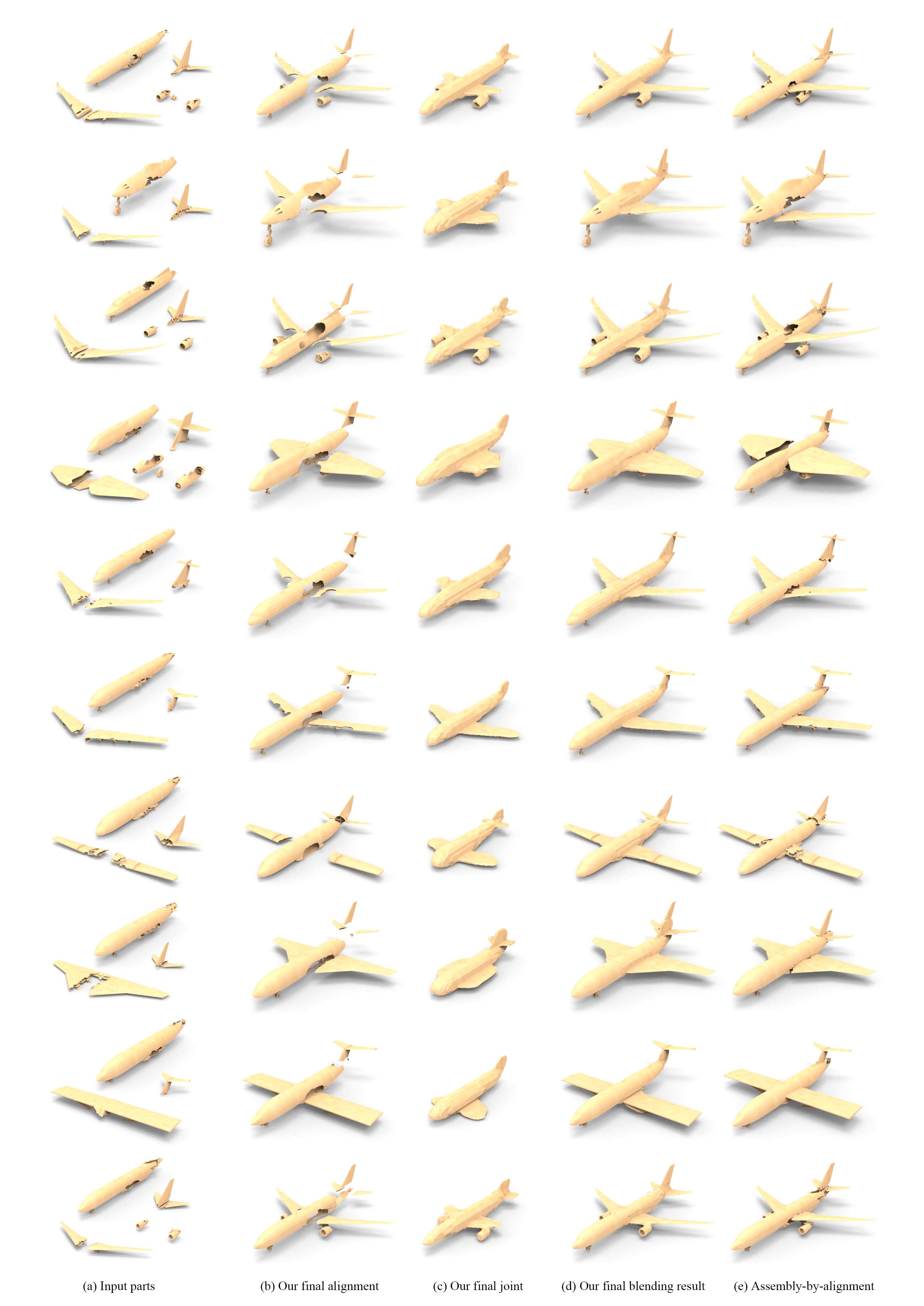}
	\caption{ More examples of airplanes.}
	\label{fig:airplane2}
\end{figure*}

We show visual results on more examples of chairs in Figure~\ref{fig:chair1},~\ref{fig:chair3},~\ref{fig:chair4}, more mugs in ~\ref{fig:mug2},~\ref{fig:mug4}, and more airplanes in Figure~\ref{fig:airplane1},~\ref{fig:airplane2}.

\subsection{Poisson mesh stitching}

\begin{figure}[b!]
	\centering
	\includegraphics[width=1.0\linewidth]{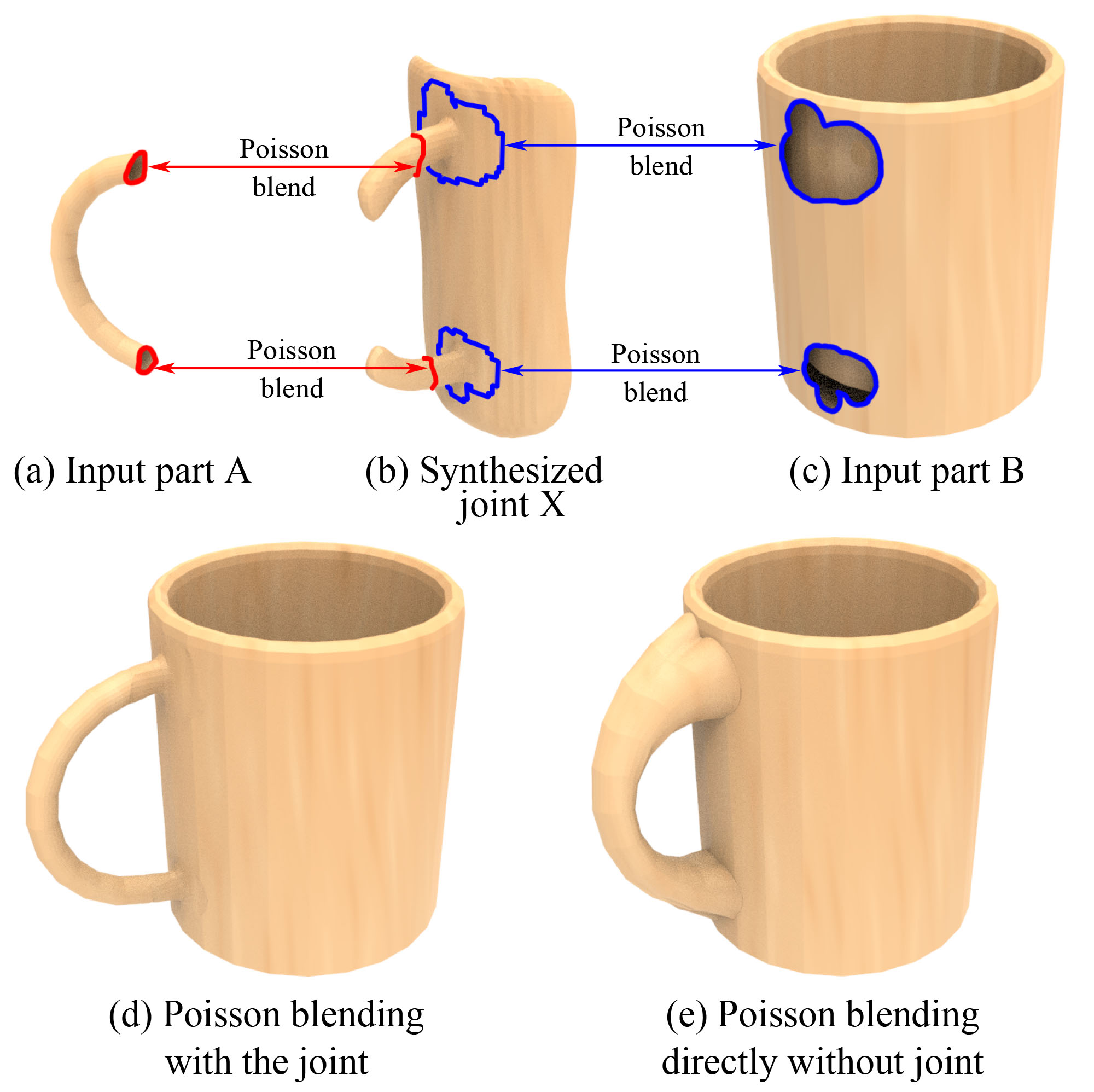}
	\caption{Poisson blending with synthesized joints. (b) shows the synthesized joint mesh for two input parts A and B. Corresponding loops are shown for the handle (red) and for the body (blue). (d) shows the output of our full pipeline which blends the parts with the synthesized joint region. (e) shows the result of naively blending the loops in A and B without the synthesized joint.}
	\label{fig:poissonblend}
\end{figure}

We use a Poisson-based approach to combine the synthesized joints with the meshes of the aligned parts to create a seamless final model. We first discretize the implicit function for the joints into a voxel grid of resolution $128^3$ and convert it into a mesh using Marching Cubes~\cite{lorensen1987marching}. The Poisson mesh stitching between this mesh and the mesh of the aligned parts has three phases: (i) finding corresponding loops between the input parts and the synthesized joints,  (ii) removing redundant regions from the joint mesh and (iii) Poisson mesh blending.

As described in the method section of the main paper, we erode the part meshes to avoid potential topological mismatches. When eroding the part meshes, the open boundaries of the eroded parts are given as directed loops as shown in Figure~\ref{fig:poissonblend} (a) and (c). To blend them with the synthesized joint mesh,  corresponding loops in the joint mesh are needed. We employ a simple approach based on nearest neighbor search to establish these correspondences as follows.

Let us denote the loop in the input part as $S$ and the loop in the synthesized joint as $T$. The joint is denoted as $X$.
The first step is to define a metric to measure the distance between a point in $S$ and a point in $X$.
We consider both euclidean distance and normal consistency and define the distance as
\[
D(\bm{p}, \bm{q}) = ||\bm{p}-\bm{q}||_2 \cdot (2 - \bm{n_p}\bm{n_q}^T)
\]
where $\bm{n_p}$ is the surface normal of point $\bm{p}$.

We randomly sample a point in $S$ as initial point, and find its closest vertex in $X$ as the first point in $T$.
Then we keep adding new points to $T$ iteratively. The algorithm is as follows.

\begin{algorithm}[]
	\SetAlgoLined
	Input: circular array $S$ that contains the vertices forming the loop in the input part\;
	Input: vertices $X$ and undirected edges $E$ of the joint\;
	Input: distance metric $D$\;
	Output: queue $T$ that contains the vertices from $X$ forming the loop in the joint\;
	Initialize $T = \O$\;
	r = a random integer\;
	initS = $S[r]$\;
	initT = the closest point to initS in $X$, according to $D$\;
	$T$.push(initT)\;
	r = r+1\;
	nextS = $S[r]$\;
	nextT = initT\;
	\While{nextS $\ne$ initS}{
		$N$ = neighbor vertices of nextT in $X$ according to $E$\;
		nextT = the closest point to nextS in $N$, according to $D$\;
		\If{nextT $\ne$ $T$.head}{
			$T$.push(nextT)\;
		}
		r = r+1\;
		nextS = $S[r]$\;
	}
	\eIf{$T$.head = $T$.tail}{
		$T$.pop()\;
		Return $T$\;
	}{
		Return None\;
	}
	\caption{Finding corresponding loop}
\end{algorithm}

Note that there is a chance that the algorithm cannot find a loop, i.e., returning ``None''.
In our implementation, we try several different initial points in $S$ to find a loop.
If they all fail, we ignore this loop and move to the next loop in the input parts.
In addition, the returned loop may contain several smaller sub-loops. In this case, we only use the largest sub-loop.

After finding corresponding loops as shown in Figure~\ref{fig:poissonblend} (b), there are redundant regions of the synthesized joint mesh which need to be removed before we can do Poisson mesh blending. Since the loops are directed, after splitting the joint mesh using the loops, each component can be marked as ``inside the part'' or ``outside the part''.  We remove the components that are detected as ``inside the part''.


After the above two steps, we use an existing Poisson mesh blending algorithm~\cite{yu2004} to merge the synthesized joint mesh with the aligned part meshes.
The Poisson blending can operate directly on the input part meshes which ensures faithful preservation of the original mesh detail without surface resampling and remeshing.
Note that after blending, there might be seams between the boundary of the joint and the boundary of the parts.
We generate a ring of triangles to fill the gap between any two corresponding loops to obtain a seamless final model.

Figure~\ref{fig:poissonblend} (e) shows the result of blending part A to part B directly without the joint. Without the joining mesh, the ends of the handle are swollen to fit the irregular open boundaries of the body. Compared to directly blending the input parts, a much better stitching result is obtained by using our synthesized joint mesh to connect the parts, as shown in Figure~\ref{fig:poissonblend} (d).

\subsection{Details of part preprocessing}

We use a part-wise point cloud representation in our neural processing. We create it by first sampling 16384 points on each shape using Poisson-disk sampling, segmenting the resulting point set into parts~\cite{yi2016scalable}, and randomly sampling $2048$ points on each part with uniform probability. Each resulting point cloud is normalized so that its centroid is at the origin and its bounding box has unit diameter. To ensure that parts can be merged together, we eliminate topological mismatches and poor segmentation boundaries by eroding the inputs around the joints. For each point, we compute the distance to the segmentation boundary and if it is less than a threshold $\tau$, the point is excluded. As an additional benefit, erosion also enables us to train the joint synthesis network in a strongly-supervised fashion, since the network has to complete the eroded regions in a training shape. Here, erosion puts the training and testing scenarios on a common footing by forcing both to process similarly eroded parts, instead of trying to synthetically simulate real-world cutting and joining errors for training.

\subsection{Details of network architecture}

In this section, we provide detailed architectures of the PointNet++'s used in our part alignment network and joint synthesis network.
Recall that in our part alignment network, we use a PointNet++ to compute the shape features for input parts. We denote this PointNet++ as PointNet++ A.
Also, in our joint synthesis network, each input part is assigned a two-branch point cloud encoder. The first branch of the part encoder is a PointNet++ network for extracting overall shape features from the parts, and we denote this PointNet++ as PointNet++ B. The second branch is another PointNet++ network for extracting localized shape features from the points that are close to the joint area, and we denote this PointNet++ as PointNet++ C.

A set abstraction layer of PointNet++ is denoted as \textit{SA}$(M,r,N,[l_1,...,l_d])$ where $M$ is the number of local patches, $r$ is the radius of balls that bound the patches, $N$ is the number of sample points selected in each patch, $[l_1,...,l_d]$ are the widths of fully-connected layers used in local PointNet.

The architectures of the three PointNet++ encoders are as follows.
\newline
\newline
PointNet++ A:
Input
$\to$ SA(256, 0.2, 128, [64, 64, 128])
$\to$ SA(128, 0.4, 128, [128, 128, 128])
$\to$ SA(1, $\inf$, $\inf$, [128, 128, 128])
$\to$ feature
\newline
\newline
PointNet++ B:
Input
$\to$ SA(256, 0.1, 128, [64, 64, 128])
$\to$ SA(128, 0.2, 128, [128, 128, 128])
$\to$ SA(1, $\inf$, $\inf$, [128, 128, 128])
$\to$ feature
\newline
\newline
PointNet++ C:
Input
$\to$ SA(256, 0.05, 128, [32, 32, 64])
$\to$ SA(128, 0.1, 128, [64, 64, 128])
$\to$ SA(1, $\inf$, $\inf$, [128, 128, 128])
$\to$ feature

\subsection{Details of network training}

The part alignment network is trained for 200 epochs with batch size 8. We use an Adam optimizer with learning rate 0.001. The training time for each data category is provided in table~\ref{table:time}.


\begin{table}[h]
	\begin{center}
	\begin{tabular}{l|l|l|l}
		\hline
		& Chair 	& Airplane & Mug \\ \hline
		Part alignment network      & 3.0    	& 2.1       & 0.1  \\ \hline
		Joint synthesis network    & 32.7     & 17.4       & 1.4  \\ \hline
	\end{tabular}

	\end{center}
	\caption{Training time (hours) on an RTX 2080 Ti GPU.}
	\label{table:time}
\end{table}

As the point cloud encoders and the IM-decoder in our joint synthesis network require different learning rates for training, we first pre-train the two-branch point cloud encoders for 100 epochs by pairing it with a fully connected decoder~\cite{achlioptas18} for reconstructing 3D shape parts. The learning rate starts at $10^{-3}$ and is halved after every 20 training epochs, until reaching $1.25 \times 10^{-4}$.  
We pair the pretrained point cloud  encoders with the IM-decoder, and train the IM-decoder with the loss function $L_{\text{joint}} = L_{\text{mse}}$ for 80 epochs, with learning rate $10^{-4}$.  To reduce training time and get more robust output, we train the IM-decoder in a  coarse-to-fine manner.  First,  the network is trained by downsampled point samples (2k samples). Then, we double the number of samples every 20 epochs, until the number reaches 32k.  Finally, we train the encoders and the IM-decoder together with the loss function $L_{\text{joint}} = L_{\text{mse}} + \alpha L_{\text{match}}$ for another 80 epochs. The default batch size is 1. The optimizer is Adam.

\begin{figure*}[h!]
	\centering
	\includegraphics[width=0.9\linewidth]{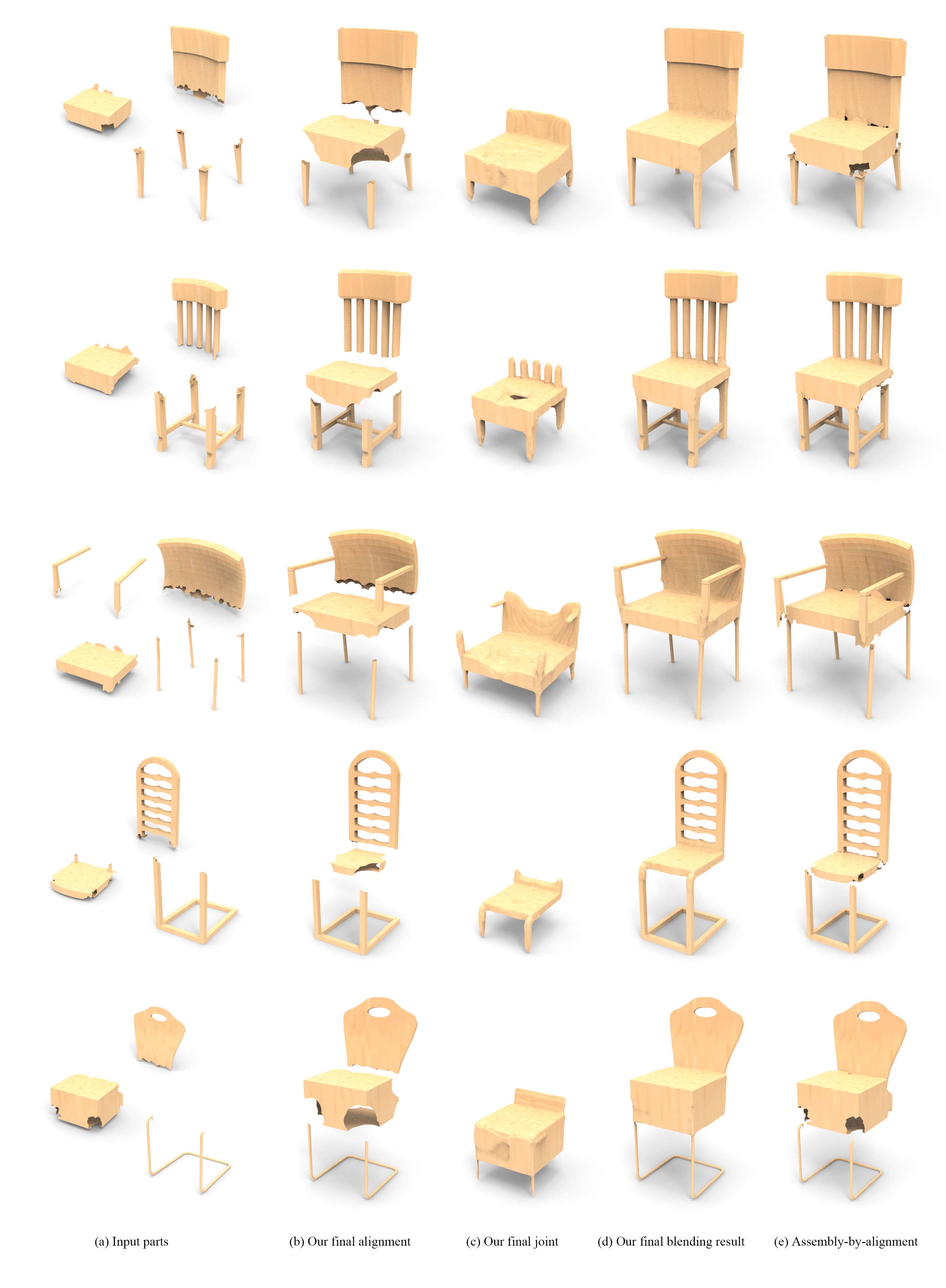}
	\caption{ More examples of chairs.}
	\label{fig:chair1}
\end{figure*}

In an early stage of our experiments, we noticed that the joint synthesis network did not work well in predicting connections for chair back with bars. After a close examination of the chair dataset, we found this is because only about 10\% of chairs have bars in their back. Therefore, we did a simple data augmentation for the chair dataset by duplicating the examples with back bars by 4 times (We did the same augmentation for all methods in our comparison experiments). After the data augmentation, our joint synthesis network works pretty well for chairs with back bars. For example, the second chair in the teaser of the paper, and the second row in Figure~\ref{fig:chair1}.

\subsection{Details of baseline networks}

We consider two voxel-based shape completion baselines that can be applied to our problem: 3DCNN and 3DResNet. In the two baselines, we use architectures that perform shape completion akin to image translation tasks. Unlike encoder-decoder architectures such as AtlasNet~\cite{Groueix18} and PCN~\cite{yuan2018pcn}, these methods do not have a bottleneck layer, and they only modify input in a relatively local manner. Thus, they do not have the challenge of reconstructing the entire shape from the latent code, and might be better at preserving the geometry of the original parts.  The 3DCNN and 3DResNet have similar architectures, but the latter uses residual skip connections. 3DCNN has 12 layers that gradually downsamples the input voxels to $16^3$ and upsamples to the output resolution; 3DResNet mimics the ResNet~\cite{he2016deep} structure for image translation used in CycleGAN~\cite{CycleGAN2017}: we simply replace 2D convolutions with 3D convolutions. The resolution of input and output voxels is $128^3$ due to the hardware (GPU memory) limitations. We train the two baseline networks on individual categories for 1 million iterations each. When testing, we use marching cubes~\cite{lorensen1987marching} to obtain the meshes from the output $128^3$ voxels.


\end{document}